\documentclass[lettersize,journal]{IEEEtran}
\usepackage{amsmath,amsfonts}
\usepackage{algorithm}
\usepackage{array}
\usepackage{subfigure}
\usepackage{textcomp}
\usepackage{stfloats}
\usepackage{url}
\usepackage{multirow}
\usepackage{verbatim}
\usepackage{graphicx}
\usepackage{booktabs}
\usepackage{cite}
\usepackage{algpseudocode}
\usepackage{setspace}
\usepackage{verbatim}
\usepackage{xcolor}

\usepackage[acronym, nohypertypes={acronym}]{glossaries}
\newacronym{model}{NexuSQN}{\emph{Nexus Sine Qua Non}}
\newacronym{task}{STTF}{Spatiotemporal Traffic Forecasting}
\newacronym{ctx}{contextualization}{contextualization}
\usepackage{multirow}
\usepackage{makecell}
\usepackage{booktabs}
\usepackage{threeparttable}

\begin{document}

\title{Contextualizing MLP-Mixers Spatiotemporally for Urban Traffic Data Forecast at Scale}

\author{Tong Nie$^\dagger$,~\IEEEmembership{Student Member,~IEEE,} Guoyang Qin$^\dagger$, Lijun Sun,~\IEEEmembership{Senior Member,~IEEE,} Wei Ma,~\IEEEmembership{Member,~IEEE,} Yu Mei, and Jian Sun$^*$
\thanks{
*This research was sponsored by the National Natural Science Foundation of China (52125208), the Science and Technology Commission of Shanghai Municipality (No. 22dz1203200), and the National Natural Science Foundation of China’s Young Scientists Fund (52302413).

Tong Nie, Guoyang Qin, and Jian Sun are with the Department of Traffic Engineering and Key Laboratory of Road and Traffic Engineering, Ministry of Education, Tongji University. Shanghai, China. 201804. (E-mail: \{nietong, 2015qgy, sunjian\}@tongji.edu.cn.)

Lijun Sun is with the Department of Civil Engineering, McGill University, Montreal, QC H3A 0C3, Canada (e-mail: lijun.sun@mcgill.ca).

Wei Ma is with the Department of Civil and Environmental Engineering, The Hong Kong Polytechnic University, Hong Kong SAR, China. E-mail: wei.w.ma@polyu.edu.hk.)

Yu Mei is with the Department of Intelligent Transportation Systems, Baidu Inc, Beijing, China. (e-mail: whqyqy@hotmail.com).

$^\dagger$ The authors contribute equally to this work.

Corresponding author: Jian Sun (sunjian@tongji.edu.cn).
}% <-this % stops a space
}

% The paper headers
\markboth{Journal of \LaTeX\ Class Files,~Vol.~14, No.~8, August~2021}%
{Shell \MakeLowercase{\textit{et al.}}: A Sample Article Using IEEEtran.cls for IEEE Journals}

% \IEEEpubid{0000--0000/00\$00.00~\copyright~2021 IEEE}
% Remember, if you use this you must call \IEEEpubidadjcol in the second
% column for its text to clear the IEEEpubid mark.

\maketitle

\begin{abstract}
Spatiotemporal traffic data (STTD) displays complex correlational structures. Extensive advanced techniques have been designed to capture these structures for effective forecasting. However, because STTD is often massive in scale, practitioners need to strike a balance between effectiveness and efficiency using computationally efficient models. An alternative paradigm based on multilayer perceptron (MLP) called MLP-Mixer has the potential for both simplicity and effectiveness. Taking inspiration from its success in other domains, we propose an adapted version, named NexuSQN, for STTD forecast at scale. We first identify the challenges faced when directly applying MLP-Mixers as series- and window-wise multivaluedness.
To distinguish between spatial and temporal patterns, the concept of ST-contextualization is then proposed. Our results surprisingly show that this simple-yet-effective solution can rival SOTA baselines when tested on several traffic benchmarks. 
Furthermore, NexuSQN has demonstrated its versatility across different domains, including energy and environment data, and has been deployed in a collaborative project with Baidu to predict congestion in megacities like Beijing and Shanghai.
% Its methodological generality is also demonstrated using data from other sources, such as energy and environment.
% Furthermore, it was deployed in a collaborative urban congestion project with Baidu, specifically evaluating its ability to forecast traffic states in megacities like Beijing and Shanghai. 
Our findings contribute to the exploration of simple-yet-effective models for real-world STTD forecasting. 
% The code is available at: \url{https://github.com/tongnie/NexuSQN}. 
\end{abstract}


\section{Introduction}
\IEEEPARstart{W}{ith} the continued growth of infrastructure and environmental systems, a vast amount of urban traffic data is being measured and collected. This data includes information on traffic flow, time stamps, and sensor locations. Analyzing urban traffic data can be challenging due to its partially observable dynamics, which include human interactions, information exchanges, and external perturbations. However, urban traffic data also exhibits macroscopic spatiotemporal correlational patterns such as continuity, periodicity, and proximity \cite{nie2023correlating}. These patterns have attracted the attention of many data-driven models that aim to capture them. 
Representative models are advanced spatial-temporal graph neural networks (STGNNs) and Transformers, which are increasingly popular and extensively studied due to their innovative designs to capture latent correlations in spatiotemporal data \cite{yu2017spatio, li2018diffusion, wu2019graph, bai2020adaptive, wu2021autoformer,zhou2022fedformer,nie2024imputeformer}.

Meanwhile, real-world urban traffic data is often massive in scale. To measure the fine-grained traffic dynamics at the network level, a large number of sensors are deployed, including both static and mobile detectors. A relatively high data recording frequency such as 5 minutes is also required. Such spatial and temporal dimensions make practical traffic data a large-scale dataset. 
For example, fine-grained highway traffic flow databases can contain more than 4 million data points \cite{liu2023largest}. Advanced modeling techniques, such as recurrent graph topology and self-attention, have high computational complexity, making them computationally intensive for city-scale applications \cite{liu2023largest,jin2023survey,cini2023graph}. Practitioners often need to strike a balance between effectiveness and efficiency by choosing simpler models and reducing complexity \cite{elsayed2021we}, particularly when resources are limited.
Trading off effectiveness for efficiency becomes a pressing question in this field. 

To break the trade-off, recent studies have brought attention back to an alternative class of simple yet effective architectures based on multilayer perceptron (MLP), called MLP-Mixers. They have demonstrated success in various domains, such as vision \cite{tolstikhin2021mlp}, language \cite{fusco2022pnlp}, and graph \cite{he2023generalization}. 
This renewed interest in MLP-Mixers suggests that they may hold promise for both effectiveness and efficiency.
{Compared to its complex counterparts such as Transformers and graph neural networks, MLP-Mixers feature lower computational complexity, thereby showing better scalability in large datasets \cite{wang2023clustering}. In addition, MLP-Mixers also show promising performances in time series forecasting tasks \cite{zeng2022transformers,zhang2023mlpst}.}
These salient features motivate us to investigate the potential of extending the MLP-Mixer paradigm to large-scale spatiotemporal urban traffic data. We aim to achieve performance comparable to elaborate architectures while maintaining simplicity and scalability.

\begin{figure*}
\centering
\includegraphics[width=1.0\textwidth]{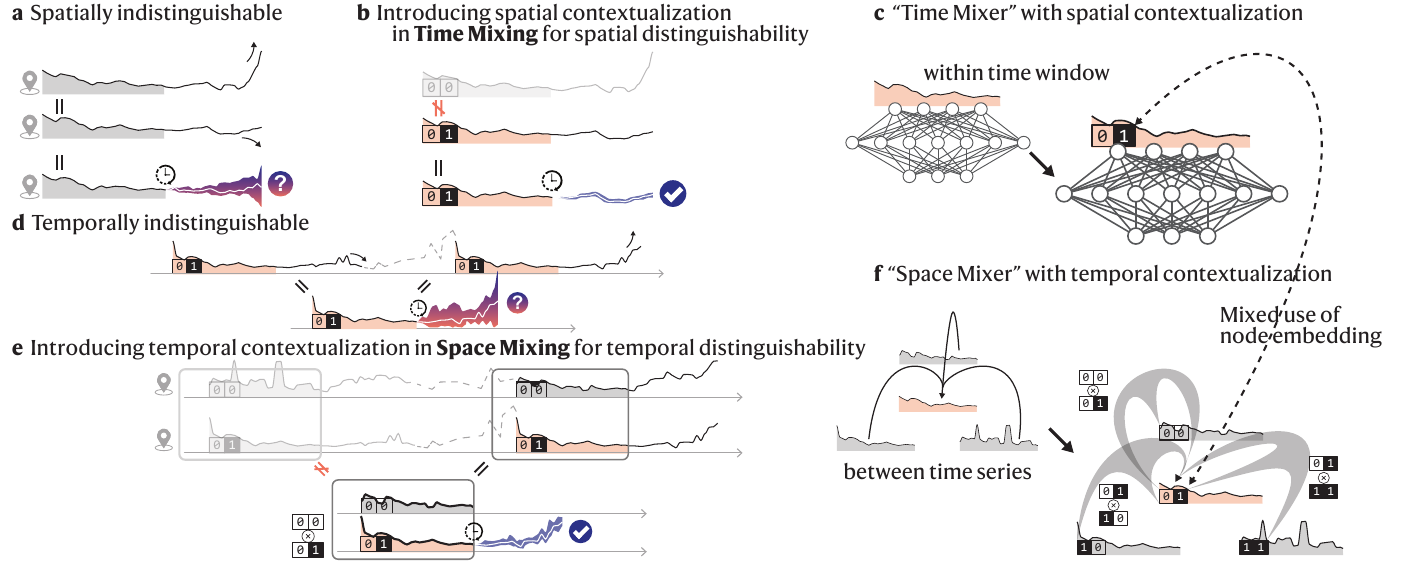}
\caption{
\textbf{ST-\acrshort{ctx} in urban traffic data}:
(a) If future series across locations differ but share a similar historical series, mixing time without knowing the location causes a series-wise multivaluedness.
(b-c) The spatial context that distinguishes the locations is needed.
(d) Even with spatial context enabled, if two windows with identical histories differ in the future, mixing space is window-wise indistinguishable. 
(e-f) The temporal context is necessary to disambiguate the multivaluedness. 
}
\label{fig:motivation}
\end{figure*}

In the context of spatiotemporal traffic data, limited pioneering work \cite{chen2023tsmixer,zhang2023mlpst,zhong2023multi,shao2022spatial,yi2023frequency} has explored the performance of MLP-Mixer on several forecasting benchmarks. However, we have found that \textit{applying it directly to a variety of real-world urban data will not yield desirable results}. After thorough investigation and surveying related work, we have identified the problem that hinders its effectiveness: the issue of \textbf{spatiotemporal contextualization}.
As illustrated in Fig. \ref{fig:motivation}, spatiotemporal contextualization refers to a data pattern where multiple possible future series share similar historical series, causing a series- and windows-wise \textbf{multivaluedness} due to the parameter sharing across locations and times.
% Multivaluedness in \acrshort{task} is identified to be twofold, one in cross-location and one within-location but cross-time-window. 
This issue is frequently posed in urban spatiotemporal systems. For instance, macroscopic traffic flow can be indistinguishable as it miss microscopic detail such as individual trajectory.
% Specifically, multivalued mappings refer to a data behavior where multiple possible future series share similar historical series and can be hard to forecast due to the absence of observations such as identity information. 
% % This often happens when the series to be forecast share traffic patterns across different times and locations. 
% This multivaluedness is ubiquitous in traffic systems, as observed macroscopic traffic flows hardly reflect microscopic processes.
When working with this multivaluedness, univalued models such as MLP-Mixers may struggle to learn nondeterministic functions. This can result in mediocre performance and systemic errors that can be hard to further reduce.
% When working with this data pattern, it can be struggling for univalued models to learn nondeterministic functions, resulting in mediocre performance and systemic errors that cannot be reduced any further. 
As a remedy, we reconstruct the contextualization information from the traffic data itself to disambiguate multivaluedness and convert them into mappings as univalued as possible. {Therefore, we expect the univalued mapping to convert the historical series, as well as the contextualization information, into a unique future series.}

By extending MLP-Mixers, in this paper, we present a scalable contextualized MLP-Mixer and investigate its application in large-scale urban traffic data forecasting problems. To address the contextualization issue, we design a learnable embedding to form a binder for spatial-temporal interactions in the mixing layers. On top of this embedding, we parameterize the time-mixer with local adaptations and customize the space-mixer with interaction-aware mixing. The spatiotemporal context makes them distinguishable by location and time. To further address the high dimensionality of large urban databases, we preserve scalability using the kernel method, achieving an efficient architecture with \textbf{linear complexity}.

Through MLP-Mixers, our focus is to explore the practical performance of a minimalist model that only includes essential connections to directly express contextualization. To evaluate the effectiveness of our model, we conduct forecasting tasks on 8 public traffic benchmarks.
Our extensive experiments, which include short-term, long-term, and large-scale forecasting tasks, reveal a surprising result -- after being spatiotemporally contextualized, MLP-Mixer can perform comparably or even surpass its more complex counterparts. Remarkably, even a single space-mixing layer can achieve competitive performance on most benchmarks.
Furthermore, MLP-Mixer offers advantages such as faster computation, lower resource consumption, a smaller model size and flexibility to transfer. Its methodological generality is also evaluated by tasks from other domains, which cover 6 urban datasets on energy consumption, meteorological records, and air quality.
% our model, which builds on a simple full MLP architecture, 

% \textcolor{purple}{[Deployed applications.]}
We also demonstrate its effectiveness in real-world deployment. We conducted an urban congestion project in collaboration with Baidu.
First, we examined its scalability to predict the propagation of large-scale congestion at the regional level in three megacities in China that contain tens of thousands of road sections. Second, we conducted online parallel tests in the generation of fine-grained congestion maps in production environments using Baidu Map's API.
% using Baidu Map's API in two urban road networks in China.
To summarize, our main technological contributions include:
\begin{enumerate}
    \item We identify the main bottlenecks of MLP-Mixers as spatiotemporal contextualization in traffic forecasting;
    \item We present a simple-yet-effective remedy to address the contextualization issue and propose a scalable model; 
    \item We showcase its superiority in accuracy, efficiency, and flexibility by comparing it to SOTA in 8 traffic benchmarks, and evaluate the methodological generality in 6 urban datasets from other sources;
    \item We deploy it in real-world large-scale urban traffic applications and demonstrate its practical effectiveness. 
\end{enumerate}
We hope this work will incentivize further endeavors to adopt and extend MLP-Mixers to more challenging urban computing problems.
As we aim to identify the essential components, we name our model \textbf{\gls{model}} networks -- a name that literally means ``\textit{the} connection without which it could \textit{not} be.''
The remainder of this paper is organized as follows. Section \ref{sec:related} provides some preliminaries and briefly reviews related work. 
Section \ref{sec:method} introduces the contextualization problem, and elaborates on our method. Section \ref{sec:exp} performs extensive experiments to evaluate it in public benchmarks. Section \ref{sec:deployed} demonstrates it real-world applications in urban network. Section \ref{sec:conclusion} concludes this work and provides future directions. The appendix provides supplementary results and discussions using urban data from other domains.

\section{Preliminary and Related Work}\label{sec:related}
\paragraph{Notations} 
% To begin, we introduce some notations using the same terminology as in previous work \cite{cini2022scalable}. 
We first introduce some notations following the terminology in \cite{cini2022scalable}. 
In a traffic network with $N$ static sensors on some locations, four types of data are considered: (1) $\mathcal{X}_{t-T+1:t}\in\mathbb{R}^{N\times T\times d_\text{in}}$: The data tensor collected by all sensors over a time interval $\mathcal{T}=\{t-T+1,\dots,t\}$, where $T$ represents the time window size, and $d_\text{in}$ is the measurement length at each sensor at each timestamp; (2) $\mathbf{A}_t\in\mathbb{R}^{N\times N}$: An adjacency matrix that can be predefined; (3) ${\mathbf{U}_{t-T+1:t} \in \mathbb{R}^{T\times d_\text{u}}}$: Exogenous variables, such as time of day information; (4) ${\mathbf{V} \in \mathbb{R}^{N\times d_\text{v}}}$: Node-specific features, such as sensor IDs. 
We denote the variables at each time step by the subscript $t$.
Note that we use the terms node, sensor, and location interchangeably. A graph signal with a window size of $T$ is defined by the following 4-tuple: $\mathcal{G}_{t-T+1:t}=\left(\mathcal{X}_{t-T+1:t}, \{\mathbf{U}_t, \mathbf{A}_t|t\in\mathcal{T}\}, \mathbf{V}\right)$. Spatiotemporal traffic forecasting (\acrshort{task}) is therefore defined as learning a parameterized function ${F(\cdot|\theta)}$ that maps to the next $H$ time steps of graph signals $\widehat{\mathcal{G}}_{t+1:t+H}$ from observation $\mathcal{G}_{t-T+1:t}$.
% , with the minimum prediction error.

% In the following sections, we will describe each module in Sec. \ref{sec:overview}. Then, we will explain the versatile use of learnable node embedding (NE) as a remedy for the \acrshort{ctx} issue in Sec. \ref{sec_loctor}. Finally, we will elaborate on our parameter-efficient design to further lighten the model in Sec. \ref{sec:mp}. Details on the model implementation are given in the appendix.

\paragraph{MLP-Mixer} Recent years have witnessed a comeback of \textsc{MLP}-based methods ({\em e.g.}, MLP-Mixer) due to its simplicity and effectiveness \cite{tolstikhin2021mlp,chen2023tsmixer}. 
% \textsc{MLP-Mixer} is a class of simple yet effective models in  \cite{tolstikhin2021mlp,he2023generalization}.
The MLP-Mixer for STTF is:
% can be formulated in a compact form as follows:
\begin{equation}
\mathbf{Y}=\textsc{MLP}_\text{time}(\textsc{MLP}_\text{channel}(\mathcal{X}_{t-T+1:t},\text{dim}=0),\text{dim}=1),
\end{equation} 
where $\textsc{MLP}_\text{time}$ and $\textsc{MLP}_\text{channel}$ alternatively perform mixing operation along temporal and spatial dimensions, respectively.
% Such a \textsc{MLP}-based architecture is shown to be competitive in time series studies \cite{zeng2022transformers,chen2023tsmixer}. 
Although recent work has already shown that the \textsc{MLP}-based architecture can be a competitive baseline for \acrshort{task} \cite{zeng2022transformers,shao2022spatial,chen2023tsmixer,zhang2023mlpst,yi2023frequency,zhong2023multi}, however, pure MLP model has been proven to be a good overfitter \cite{he2023generalization} and easily falls into undesirable local suboptimal solutions. In addition, directly applying them to various \acrshort{task} tasks cannot achieve competitive performances due to the contextualization issue. 
% To address this issue, we propose exploiting the relational structures to increase the expressivity of \textsc{MLP-Mixer}.

\paragraph{Related Work} As the leading approaches for \acrshort{task}, STGNNs and Transformers use advanced techniques such as GNNs, RNNs and self-attention to exploit spatiotemporal correlations for effective forecasting. Extensive studies have shown promising results using them \cite{yu2017spatio, li2018diffusion, wu2019graph, bai2020adaptive, guo2019attention, zheng2020gman, wu2021autoformer,zhou2022fedformer,nie2022time,he2024geolocation}. Although existing architectures perform well, they can require a high complexity to implement. Previous research has attempted to address this scalability issue by simplifying the architecture or using better embedding techniques.

Among the limited studies on simplified and scalable methods for \acrshort{task}, \cite{cini2022scalable} proposed scalable STGNNs based on preprocessing that encode data features prior to training. \cite{liu2023we} replaced the GNNs with alternative spatial techniques and used a graph sampling strategy to improve performance. Both of the methods elaborate temporal encoders and pre-computed graph features. 
% In particular, \cite{cini2022scalable} proposed a scalable graph predictor based on the random-walk diffusion operation and the echo-state network to encode spatiotemporal representations prior to model training. 
% \cite{liu2023we} developed two alternative spatial techniques including a pre-processing-based ego graph method and a global sensor embedding to model spatial correlations. The processed spatial features are further fed to temporal models such as RNNs, TCNs, and WaveNets. A graph sampling strategy is required to improve training performance. However, both of the two approaches rely on complex temporal encoders and precomputed graph features. 
% \cite{satorras2022multivariate} developed a fully connected gated GNN model using a graph inference structure. However, pairwise attention can incur high computational costs. 
TS-Mixer \cite{chen2023tsmixer}, MLPST \cite{zhang2023mlpst}, FreTS \cite{yi2023frequency}, ST-MLP \cite{wang2023st}, PITS \cite{lee2023learning} and MSD-Mixer \cite{zhong2023multi} studied the effectiveness of MLP-based architectures in medium-sized forecasting benchmarks and showed promising performance.
In another vein of research, several methods have emerged to challenge Transformer-based models existing for long-term series forecasting (LTSF), such as LightTS \cite{zhang2022less}, DLinear \cite{zeng2022transformers}, and TiDE \cite{das2023long}. Most of them adopt a channel independence strategy \cite{nie2024channel} and do not include explicit relational modeling.

Regarding embedding methods for spatiotemporal data, \cite{shao2022spatial} used learnable embeddings for all nodes, time of day, and day of week points. This approach enables MLPs to achieve competitive performance on several datasets by learning identity information. However, using overparameterized embedding can lead to redundancy and overfitting. 
% Additionally, temporal identifiers of this type only reflect time-varying components and are less effective in capturing local spatial dynamics. 
\cite{cini2023taming} offered further interpretation of node embedding as local effects and integrated it into a global-local architecture. Fourier positional encoding was adopted in \cite{nie2024spatiotemporal} to enable canonical MLPs to capture high-frequency patterns in spatiotemporal data. Geolocation encodings generated by pretrained language models are adopted in \cite{he2024geolocation} to enhance spatiotemporal forecasting.

% \textcolor{gray}{In summary, previous work did not directly investigate or thoroughly discuss the essential model components for \acrshort{task} or contextualization issues in STGNNs, leaving opportunities for this study to contribute.}
In summary, STGNNs and Transformers have been extensively studied in \acrshort{task},
but MLP-Mixers for large-scale urban traffic data have been less investigated, leaving opportunities for us to contribute.

\section{ST-Contextualized MLP-Mixers}\label{sec:method}

To address the contextualization issue, this section demonstrates the proposed \gls{model} model. 
% Our model is based on a minimalistic architecture that employs a scalable message-passing (MP) layer with ``where'' and ``when'' locators, as shown in Fig. \ref{fig:arch}. 
\gls{model} is a simple yet effective architecture that features a full MLP-based structure with spatial-temporal interactions, as shown in Fig. \ref{fig:motivation}. 
Notably, it leverages the structured mixing operations to contextualize the \acrshort{task} task, allowing it to bypass the need for complex temporal methods (e.g., RNN, TCN, and self-attention) and spatial techniques (e.g., diffusion convolution and predefined graph). {It also has \textbf{linear complexity} with respect to the length of the sequence and the number of series: given $N$ traffic time series with a window length $T$, the time complexity of a forward computation of the model scales linearly, i.e. $\mathcal{O}(N+T)$.}

\subsection{Overview of Architectural Components}\label{sec:overview}
% Model Structure 
% \textcolor{brown}{Plz unify the notation throughout the paper!!!}

\paragraph{Overview} The overall architecture of \gls{model} can be concisely formulated as follows:
\begin{equation}\label{model_structure}
\begin{aligned}
    &\mathbf{H}^{(1)}=\textsc{TimeMixer}(\mathcal{X}_{t-T+1:t};\mathbf{E}_{t-T+1:t}),\\
    &\mathbf{H}^{(l+1)}=\textsc{SpaceMixer}(\mathbf{H}^{(l)};\mathbf{E}_{t-T+1:t}),~l\in\{1,\dots,L\},\\
    &\mathcal{X}_{t+1:t+H}=\textsc{Readout}(\mathbf{H}^{(L+1)}),\\
\end{aligned}
\end{equation}
% \begin{equation}\label{model_structure}
% \begin{aligned}
%     &\mathbf{H}^0=\textsc{Projection}(\mathcal{X}_{T-W+1:T}),\\
%     &\mathbf{H}^{(1)}=\textsc{TimeMixer}(\mathbf{H}^{(0)};\mathbf{E}_{T-W+1:T}),\\
%     &\mathbf{H}^{(l+1)}=\textsc{SpaceMixer}(\mathbf{H}^{(l)};\mathbf{E}_{T-W+1:T}),~l\in\{1,\dots,L\},\\
%     &\mathcal{X}_{T+1:T+H}=\textsc{Readout}(\mathbf{H}^{(L+1)}),\\
% \end{aligned}
% \end{equation}
% where \textsc{TimeMixer} and \textsc{SpaceMixer} build on conceptually and technically simple \textsc{Mlp} and \textsc{MPNN} blocks, which can be represented together by Fig. \ref{fig:arch}(b). $\mathbf{E}$ is the learnable NE, and $\mathbf{A}$ is the relational graph. In the subsequent paragraphs, we will detail each component in Eq. \ref{model_structure} in turn.
\noindent where the \textsc{TimeMixer} and \textsc{SpaceMixer} modules are extending the conceptually simple MLP-Mixers to consider spatiotemporal contextualization. Importantly, $\mathbf{E}_{t-T+1:t}$ represents the spatiotemporal node embedding (STNE). 
% In the following paragraphs, we will elaborate on each component in Eq. \eqref{model_structure}. 
\textsc{Readout} module contains a $\textsc{Mlp}$ and a reshaping layer to directly output multistep predictions. As shown, only one time-mixing operation is performed, followed by $L$ space-mixing layers. Interestingly, we find that a single \textsc{SpaceMixer} can achieve competitive performance on most benchmarks.

% \paragraph{Dense \normalfont\textsc{Readout}}

% where $\texttt{UNFOLD}(\cdot)$ is the inverse linear operator of $\texttt{FOLD}(\cdot)$. 
% Again, we do not elaborate on a complex sequential decoder, and the multistep predictions are regressed directly.

\paragraph{{Spatiotemporal Node Embedding}} 
To address the \acrshort{ctx} issue, we use the versatility of node embedding as a structural cornerstone and enable it to form a binder for spatial-temporal interactions in the mixing layers.
It mimics the positional and structural representations in the theory of GNNs' expressivity \cite{srinivasan2019equivalence}. 
% The roles of node embedding in our model are twofold: a ``\emph{where locator}'' and ``\emph{when locator}.'' 
% They serve as positional and structural representation in the theory of GNNs' expressivity \cite{srinivasan2019equivalence}. 
% \textcolor{purple}{ And can this STNE be formulated with a factorization/multiplication?}
% Regarding implementation, we assign a learnable vector of size $d_{\text{emb}}$ with random initialization for each series, denoted as $\mathbf{E}\in\mathbb{R}^{N\times d_{\text{emb}}}$, and then include $\mathbf{E}$ in the forward process as endogenous variables. Its gradient is updated end-to-end by backpropagation.
Learnable node embedding has been adopted as a positional encoding in both spatiotemporal and general graphs \cite{shao2022spatial,dwivedi2022graph,cini2023taming}. Given $N$ nodes, it can be instantiated as a random dictionary $\mathbf{E}\in\mathbb{R}^{N\times d_{\text{emb}}}$. However, it only reflects the static features of each series, such as the dominating patterns, and is agnostic to temporal variation. 
To address this, we propose incorporating the stationary property of time series, such as periodicity and seasonality.
We use sinusoidal positional encoding \cite{vaswani2017attention} $\mathbf{U}_{t-T+1:t}\in\mathbb{R}^{T\times d_\text{u}}$ to inform $\mathbf{E}$ with time-of-day context. 
Specifically, we first project the two encodings to a matched dimension and fuse them with a MLP:

% This NE reflects the static features of each sensor, e.g., dominating traffic patterns \cite{nie2023correlating}, and is agnostic to temporal information. Inspired by \cite{marisca2022learning}, we propose to inform the model with stationary serial property of traffic flows, e.g., periodicity and seasonality. The sinusoidal positional encoding \cite{vaswani2017attention} $\mathbf{U}_t\in\mathbb{R}^{T\times d_s}$ is adopted to inject the time-of-day information to all series. Given $\mathbf{U}_t$, we first fold the time dimension and project it into the hidden space:
% \begin{equation}\label{STNE}
%     \begin{aligned}
%     \widetilde{\mathbf{u}}&=\mathbf{W}_\text{U}\texttt{FOLD}(\mathbf{U}_t),\\
%     \widetilde{\mathbf{U}}&=\texttt{BROADCAST}(\widetilde{\mathbf{u}},N),\\
%     \mathbf{E}_t&=\textsc{ResMlp}(\textsc{ResMlp}(\mathbf{E}+\widetilde{\mathbf{U}})),
%     \end{aligned}
% \end{equation}
% \begin{equation}\label{STNE}
%     \begin{aligned}
%     \widetilde{\mathbf{u}}&=\mathbf{W}_\text{U}\texttt{FOLD}(\mathbf{U}_t),\\
%     \widetilde{\mathbf{U}}&=\texttt{BROADCAST}(\widetilde{\mathbf{u}},N),\\
%     \mathbf{E}_t&=\textsc{Mlp}([\mathbf{E}\|\widetilde{\mathbf{U}}]),
%     \end{aligned}
% \end{equation}

\begin{equation}\label{STNE}
    \begin{aligned}
    &\widetilde{\mathbf{U}}=\mathbf{W}_\text{u}\mathbf{U}_{t-T+1:t}, ~\widetilde{\mathbf{E}}=\mathbf{E}\mathbf{W}_\text{e},\\
    &\mathbf{E}_{t-T+1:t}=\textsc{Mlp}(\widetilde{\mathbf{E}}\widetilde{\mathbf{U}}^{\mathsf{T}}),
    \end{aligned}
\end{equation}
\noindent where $\mathbf{W}_\text{u}\in\mathbb{R}^{D\times T}$ and $\mathbf{W}_\text{e}\in\mathbb{R}^{d_{\text{emb}}\times d_{\text{u}}}$
are weights, 
% and $\texttt{BROADCAST}(\cdot)$ repeats the inputs with given times along a new dimension. 
and $\mathbf{E}_{t-T+1:t}\in\mathbb{R}^{N\times D}$ is the STNE. When inputs from different intervals exhibit varying patterns, STNE is able to capture the variability of series that creates a dynamic representation.
% The following paragraphs explain how STNE contributes to forecasting.

\subsection{Flattened Dense Time-Mixers}

% \paragraph{Input Flattening and \normalfont\textsc{Projection}}\label{s:encoder} 
\paragraph{Dimension Flattening and Projection}\label{s:encoder} 
STGNNs usually handle time and feature dimensions separately. This involves first independently projecting the input features at each time step to high-dimensional features and then correlating different time steps with sequential models such as RNNs. While this treatment is reasonable and intuitive, it significantly increases model complexity and the risk of overfitting, especially when the hidden size is much larger than the feature dimension. To address this issue, we propose to flatten the input series along the time dimension and project it into the high-dimensional hidden space using a single MLP:
% The time and feature dimensions are usually handled separately in STGNNs, i.e., first projecting the input features at each time step to high-dimensional representations independently and then correlating different time steps with sequential models such as RNNs. Although reasonable and intuitive, this treatment increases the model complexity and the risk of overfitting significantly, especially when the hidden size is much larger than the channel (feature) dimension. As such, we propose to flatten raw time series along time dimension and project the inputs into hidden space with a simple $\textsc{Mlp}$ layer:
% \begin{equation}
% \begin{aligned}
%     &\mathbf{X}_{T-W+1:T} = \texttt{FOLD}([\mathcal{X}_{T-W+1:T}\|\mathbf{U}_{T-W+1:T}]),\\
%     &\mathbf{H}^0 =\textsc{Mlp}(\mathbf{X}_{T-W+1:T}),\\
%     % &\mathbf{H}^1 =\textsc{ResMlp}(\mathbf{H}^0 + \mathbf{E}_t),\\
% \end{aligned}
% \end{equation}
\begin{equation}\label{input_flat}
\begin{aligned}
    &\mathbf{X}_{t-T+1:t} = \texttt{FOLD}(\mathcal{X}_{t-T+1:t}),\\
    &\mathbf{H}^{(0)} =\textsc{Mlp}(\mathbf{X}_{t-T+1:t}),\\
    % &\mathbf{H}^1 =\textsc{ResMlp}(\mathbf{H}^{(0)} + \mathbf{E}_t),\\
\end{aligned}
\end{equation}
where $\texttt{FOLD}(\cdot):\mathbb{R}^{N\times T\times d_{\text{in}}}\rightarrow\mathbb{R}^{N\times T d_{\text{in}}}$ represents the flattening operator. By flattening and projecting along the time axis, serial information is stored in the hidden state $\mathbf{H}^{(0)}\in\mathbb{R}^{N\times D}$.
% In addition, we incorporate exogenous variables $\mathbf{U}_{T-W+1:T}$ to provide the projection layer with the property of time series, such as periodicity and seasonality.

% \paragraph{TimeMixer for Exploiting Temporal Relations} 
\paragraph{Time Mixing with Spatial Context} 
To further encode the temporal relations and patterns underlying the historical series, \textsc{TimeMixer} further adopts a two-layer feedforward $\textsc{Mlp}$ for time mixing:
% Time mixer models the temporal relations and patterns underlying historical series. We adopt a 2-layer $\textsc{Mlp}$ with residual connection to encode temporal representations of traffic flows: 
% \begin{equation}\label{time_mixer}
% \begin{aligned}
%     \widetilde{\mathbf{H}}^0 &=\texttt{GeLU}(\mathbf{H}^0\mathbf{\Theta}_{\text{time}}^0),\\
%     \mathbf{H}^1 &=\texttt{LayerNorm}(\widetilde{\mathbf{H}}^0)+\mathbf{H}^0\mathbf{\Theta}_{\text{time}}^1,\\
% \end{aligned}
% \end{equation}
\begin{equation}\label{time_mixer}
\begin{aligned}
    \widetilde{\mathbf{H}}^{(0)} &=\textsc{Mlp}(\mathbf{H}^{(0)}),\\
    \mathbf{H}^{(1)} &=\texttt{LayerNorm}(\widetilde{\mathbf{H}}^0+\textsc{MLP}(\mathbf{H}^{(0)})),\\
\end{aligned}
\end{equation}
% \noindent where $\mathbf{\Theta}_\text{time}^0$ and $\mathbf{\Theta}_\text{time}^1$ are linear weights, 
where $\mathbf{H}^{(1)}$ is the time-mixed representation,
and $\texttt{LayerNorm}(\cdot)$ is adopted to reduce the variance between multivariate series \cite{liu2023itransformer}.
% mitigate the distribution shift \cite{kim2021reversible} and reinforce the representations of high-frequency temporal components \cite{deng2021st}.
% where $\mathbf{\Theta}_{\text{time}}^0,\mathbf{\Theta}_{\text{time}}^1$ are linear weights, $\texttt{IN}(\cdot)$ is the instance normalization \cite{ulyanov2016instance}, which is adopted to mitigate distribution shift \cite{kim2021reversible,nie2022time} and reinforce the representations of high-frequency temporal components \cite{deng2021st}. 
% \paragraph{``Where Locator''}
% \paragraph{\textcolor{purple}{Series-wise multivaluedness.}} 
It is observed that Eqs. \eqref{time_mixer} and \eqref{input_flat} are global models \textbf{shared by all series}. Given two sensors with similar historical input but different dynamics in the future, these global models can fail to contextualize each series in spatial dimension, causing a \textit{series-wise multivaluedness}. A straightforward way to adapt the global model to each series is to set a specialized encoder for each one, that is:
\begin{equation}
    \widetilde{\mathbf{h}}_i^{(0)} =\textsc{Mlp}_i(\mathbf{h}_i^{(0)}), \forall i=\{1,\dots,N\}.
\end{equation}
However, such an overparameterization is undesirable and prone to overfitting. To address this issue, we customize each node with a discriminative identifier using STNE directly. Specifically, we concatenate $\mathbf{E}_{t-T+1:t}$ to all series in the first \textsc{Mlp} of \textsc{TimeMixer} in Eq. \eqref{time_mixer} as spatial context for distinguishable time mixing:
% It can be seen that Eqs. \ref{time_mixer} and \ref{graph_mixer} are global models shared by all sensors. Given two sensors with similar historical input, global ones fail to contextualize each series in the spatial dimension, even though they will have different dynamics in the future. To circumvent such a one-to-many problem, we parameterize each node with a discriminative identifier using STNE directly. Especially, we add the learnable $\mathbf{E}_t$ to each series in the first \textsc{Mlp} of \textsc{TimeMixer} in Eq. \ref{time_mixer} as spatial \acrshort{ctx}:
\begin{equation}
    \widetilde{\mathbf{h}}_i^{(0)} =\textsc{Mlp}([\mathbf{h}_i^{(0)}\|\mathbf{e}_i^t]), \forall i=\{1,\dots,N\},
\end{equation}
% \textcolor{purple}{(Can there be another paragraph to formally describe the local components?)
which yields:
% \begin{equation}
$[\mathbf{h}_i^{(0)}\|\mathbf{e}_i^t]\mathbf{\Theta}=[\mathbf{h}_i^{(0)}\|\mathbf{e}_i^t]\left[\begin{aligned}
    \mathbf{\Theta}_1 \\
    \mathbf{\Theta}_2
\end{aligned}\right]=\mathbf{h}_i^{(0)}\mathbf{\Theta}_1+\mathbf{e}_i^t\mathbf{\Theta}_2$,
% \end{equation}
where $\mathbf{\Theta}_1$ is the global component shared by all nodes, while $\mathbf{e}_i\mathbf{\Theta}_2$ is the specialized local adaptation of individuals.

The introduction of learnable embeddings for each node is equivalent to the specialization of the global model, which improves the use of local dynamics specific to nodes \cite{cini2023taming}. We also find that adding STNE to each series can be treated as incorporating low-frequency components into high-frequency representations. This highlights the impact of local events on urban spatiotemporal traffic data. 
% In fact, random node initialization itself improves GNNs \cite{sato2021random}. 
% Note that ``where locator'' only implies the exclusivity of the identifier, and we set it to be learnable for the sake of reuse for ``when locator.''

\subsection{Structured Scalable Space-Mixers}
To model the multivariate correlations of series, the canonical \textsc{MLP}-Mixers for space mixing is:
\begin{equation}
\begin{aligned}
&\bar{\mathbf{h}}^{d}=\sigma(\mathbf{\Theta}_{\text{channel}}\mathbf{h}^{d}+\mathbf{b}),\\
    \text{or:~}&\bar{h}_{n}=\sigma(\sum_{j=1}^N \theta_{n,j}h_{j}+b_{n}),~n\in\{1,\dots,N'\},
\end{aligned}
\end{equation}
where $\bar{\mathbf{h}}^{d}\in\mathbb{R}^{N'}$ is the graph state at dimension $d$, and $\mathbf{\Theta}_{\text{channel}}\in\mathbb{R}^{N'\times N}$ is channel mixing parameter. The unconstrained mixing weight $\theta_{n,j}$ is \textbf{shared by all time windows} and is agnostic to the absolute position in the sequence, rendering it incapable of contextualizing the series in temporal dimension, causing a \textit{window-wise multivaluedness}. For instance, a sensor's readings may differ at two different windows but share a similar historical series. In addition, the mixed message from $h_{j}$ depends only on the state of node $j$ and is agnostic to the temporal pattern of $h_{n}$. The interaction between two temporal patterns should also be considered. 

% Although STNE can be understood as a global identifier, relying solely on the ``where locator'' may be ineffective in contextualizing the temporal location of a series. 
% A natural way to remedy drawbacks of time-dependent models is to assign a unique identifier to each time point. However, this approach has two issues: 
A remedy for this issue is to assign a unique identifier to each time point. However, this approach has two issues: 
% (1) having different identifiers within a window is redundant since the prediction is basically done sequence-to-sequence; 
(1) time independent model relies on relative timestamps and using absolute ones is difficult for it; (2) timestamp-based encoding can attribute the ambiguity to static factors, such as time of day, but this could be caused by time-varying dynamics and specific temporal patterns.

% \textcolor{purple}{however, the mixing weight $\mathbf{W}_i$ is agnostic to the specific time window.}

% In the discussion by \cite{chen2023tsmixer}, predictive models like Eq. \eqref{autoreg} are referred to as time-dependent. 

\paragraph{Space Mixing with Temporal Context} To enable the \textsc{SpaceMixer} to become time-varying and adaptive to the interactive temporal context within time windows,
% To make the model aware of the temporal variation patterns and time stamp information, an upgrade to this type is termed data-dependent, 
we consider a pattern-aware mixing:
\begin{equation}
\bar{h}_{n}=\sigma(\sum_{j=1}^N \mathcal{F}_t(\mathbf{h}^d)h_{j}+b_{n}),~n\in\{1,\dots,N'\},
\end{equation}
where $\mathcal{F}_t(\cdot)$ represents a data-dependent function, such as self-attention. 
% Eq. \eqref{time_varying_ar} creates a fully time-varying AR process, which is parameterized by time-varying coefficients \cite{bringmann2017changing}. 
% However, its overparameterization can lead to overfitting of the data, rather than capturing the temporal relationships, such as the position on the time axis. Thus, new methods are required to design a ``when locator.''
To implement, we set the mixed dimension $N'=N$ to preserve the spatial dimension and formulate a structured \textsc{SpaceMixer}:
% \begin{equation}\label{MPNN}
% \begin{aligned}
%     &\mathbf{m}_t^{i\leftarrow j,(l)}=\textsc{Msg}_l(\mathbf{h}_i^{(l-1)},\mathbf{h}_j^{(l-1)})=\Psi(\mathbf{h}_i^{(l-1)}\|\mathbf{h}_j^{(l-1)})\mathbf{h}_j^{(l-1)},\\
%     &\mathbf{m}_t^{i,(l)}=\textsc{Agg}(\mathbf{m}_t^{i\leftarrow j,(l)};\forall j\in\mathcal{N}(i))=\sum_{j\in\mathcal{N}(i)}\mathbf{m}_t^{i\leftarrow j,(l)},\\
%     &\mathbf{h}_t^{i,(l)}=\textsc{Up}_l(\mathbf{h}_t^{i,(l-1)},\mathbf{m}_t^{i,(l)})=\sigma(\mathbf{\theta}^{\mathsf{T}}\mathbf{h}_t^{i,(l-1)}+\mathbf{m}_t^{i,(l)}),\\
% \end{aligned}
% \end{equation}
\begin{equation}\label{MPNN}
\begin{aligned}
    &\mathbf{m}_{i\leftrightarrow j}^{(l+1)}=\textsc{Contx}_l(\mathbf{h}_i^{(l)},\mathbf{h}_j^{(l)})=\Psi(\mathbf{h}_i^{(l)}\|\mathbf{h}_j^{(l)})[\mathbf{h}_i^{(l)}\|\mathbf{h}_j^{(l)}],\\
    &\mathbf{m}_i^{(l+1)}=\textsc{Mix}(\mathbf{m}_{i\leftrightarrow j}^{(l+1)};\forall j=\{1,\dots,N\})=\sum_{j=1}^{N}\mathbf{m}_{i\leftrightarrow j}^{(l+1)},\\
    &\mathbf{h}_i^{(l+1)}=\textsc{Update}_l(\mathbf{h}_i^{(l)},\mathbf{m}_i^{(l+1)})=\sigma(\mathbf{\Theta}\mathbf{h}_i^{(l)}+\mathbf{m}_i^{(l+1)}),\\
\end{aligned}
\end{equation}
where $\mathbf{h}_i^{(l)}$ is the representation of node $i$ in layer $l$, and $\mathbf{\Theta}$ is the feedforward weight. $\Psi$ is the temporal context function.

Compared to MLP-Mixer $\mathbf{H}^{(l)}=\sigma(\mathbf{\Theta}^{(l)}_\text{channel}\mathbf{H}^{(l-1)}\mathbf{\Theta}^{(l)}_\text{time})$, Eq. \eqref{MPNN} customizes the channel mixing weight $\mathbf{\Theta}_\text{channel}$ with structured function $\Psi(\mathbf{h}_i^{(l-1)}\|\mathbf{h}_j^{(l-1)})$. Note that the contextualization function should be specified to be aware of the specific temporal interactions and positions at current window, $\Psi$ requires the pairwise correlating of node features. 
% Essentially, we only need to specify the global message function $\Psi(\mathbf{h}_t^{i,(l-1)}\|\mathbf{h}_t^{j,(l-1)})$ in Eq. \eqref{MPNN} using the ``where locator'' directly. 
However, such a global operation costs $\mathcal{O}(N^2)$ complexity, making it inefficient for large-scale networks. To address this scalability issue, we propose a scalable kernelized mixing using the STNE directly to reparameterize Eq. \eqref{MPNN}. 

\paragraph{Scalable Kernelized Contextualization} Recall that $\mathbf{E}_{t-T+1:t}$ not only includes time information but can also be used as an agent to measure interactions between different series. To this end, we suggest \textbf{reusing} the STNE as a query for all available references to contextualize temporal patterns. 
% To design an effective and efficient ``when locator,'' we eventually propose to exploit rich information of other nodes to contextualize the current series. Given that the temporal dynamics of other series provides referable information about the specified period, it can itself be used as a ``when locator.'' 
% \textsc{Mpnn} can be inadequate to distinguish two nodes with a similar neighborhood structure. This issue is known as the spatial indistinguishability problem \cite{dwivedi2022graph}. 
% To allow the message function to be \textbf{aware of the identity (local patterns) of both sender and receiver nodes} and preserve discriminative features even after message aggregation, we adopt anisotropic message functions in \textsc{Msg} \cite{cini2023taming}. 
% In addition, stacking layers of \textsc{Mpnns} or using a high-order \textsc{Mpnn} layer can alleviate the spatial contextualization issue. However, these treatments are prone to the over-squashing problem \cite{dwivedi2022graph} and deficient in temporal contextualization.
% Remedies for this issue are discussed in Sec. \ref{sec_loctor}.
% Considering the dependent dynamics of channels (sensors), we can formulate the \textsc{Mlp} mixer as:
Specifically, the temporal query $\langle\mathbf{e}_i^t,\mathbf{e}_j^t\rangle$ can be treated as the vector inner product in the Hilbert space $\mathcal{H}$. Define a kernel function $\mathcal{K}: \mathcal{X}\times \mathcal{X}\rightarrow \mathbb{R}$, it admits:
% the following property:
\begin{equation}
    \mathcal{K}(\mathbf{e}_i^t,\mathbf{e}_j^t)=\langle\phi(\mathbf{e}_i^t),\phi(\mathbf{e}_j^t)\rangle_\mathcal{H},
\end{equation}
where $\phi: \mathcal{X}\rightarrow \mathbb{H}$ is a mapping function. Typically, the kernel method directly defines the kernel function and bypasses the explicit form of $\phi$. Instead, if we approximate the contextualization function $\Psi(\mathbf{h}_i^{(l-1)}\|\mathbf{h}_j^{(l-1)})$ using the kernel definition, we have:
\begin{equation}
\begin{aligned}
    ({\Psi}\mathbf{H})_i &= \frac{\sum_{j=1}^N\mathcal{K}(\mathbf{e}_i^t,\mathbf{e}_j^t)\mathbf{h}_j}{\sum_{j=1}^N\mathcal{K}(\mathbf{e}_i^t,\mathbf{e}_j^t)}
    = \frac{\sum_{j=1}^N (\phi(\mathbf{e}_i^{t,\mathsf{T}})\cdot\phi(\mathbf{e}_j^t))\mathbf{h}_j^{\mathsf{T}}}{\sum_{j=1}^N\phi(\mathbf{e}_i^{t,\mathsf{T}})\cdot\phi(\mathbf{e}_j^t)}, \\
\end{aligned}
\end{equation}
where the conditional probability $p(\mathbf{e}_j^t|\mathbf{e}_i^t)=\frac{\mathcal{K}(\mathbf{e}_i^t,\mathbf{e}_j^t)}{\sum_{j=1}^N\mathcal{K}(\mathbf{e}_i^t,\mathbf{e}_j^t)}$ is a linearized kernel smoother \cite{tsai2019transformer}.
The above process can be simplified using the associative property of matrix multiplication:
\begin{equation}\label{kernel_mp}
    ({\Psi}\mathbf{H})_i = \frac{\phi(\mathbf{e}_i^{t,\mathsf{T}})\sum_{j=1}^N (\phi(\mathbf{e}_j^t)\mathbf{h}_j^{\mathsf{T}})}{\phi(\mathbf{e}_i^{t,\mathsf{T}})\sum_{j=1}^N\phi(\mathbf{e}_j^t)}.
\end{equation}

In this way, we can compute the above equation in \textbf{$\mathcal{O}(N)$ time and space complexity} with a proper kernel function. Using Eq.\eqref{kernel_mp}, \textsc{SpaceMixer} can efficiently capture pairwise interactions among windows as the temporal context for distinguishable space mixing.
% The choice of proper kernel functions will be discussed in section xxx.

Since $\mathbf{E}_t$ contains the time-of-day feature, this query function can vary over time. This makes our approach lie between time-dependent and data-dependent routines, allowing it to forecast using both relative timestamps and interactive temporal patterns. Furthermore, it models the positional embeddings and structural representations simultaneously in spatiotemporal graphs \cite{srinivasan2019equivalence}.

\subsection{Remarks on Parameter-Efficient Designs}\label{sec:mp}
The proposed scalable spatiotemporal mixing layers also have several parameter-efficient designs. As \gls{model} conducts space mixing only after one time mixing rather than at every time point, it is more efficient than alternately stacking spatial-temporal blocks. We introduce these techniques and explain how they lighten the model structure. 
% \begin{enumerate}
%     \item Dense graph with fewer layers: we use small number of dense graph aggregations (e.g., 1 layers in most cases) to capture large spatial receptive field;
%     % \item Shared mixing layer: the \texttt{linear} operator is shared between consecutive MP layers to mix the time dimensions.
% \end{enumerate}

% A key innovation of \gls{model} is its parameter-efficient message passing backbone. 
% We introduce several task-agnostic techniques to lighten the model structure. 
% Another ingenuity of \gls{model} is its parameter-efficient MP backbone. We present several task-agnostic techniques to lighten the overall neural architecture. Note that since \gls{model} propagates features only after the time mixing step, rather than at every time point, it is more efficient than alternately stacked spatiotemporal blocks.

\paragraph{Residual Connection}
In each block, we incorporate a shortcut for the linear part. When all residual connections are activated (e.g., when spatial modeling is unnecessary), \gls{model} can degrade into a family of channel-independent linear models. This feature makes it promising for long-term series forecasting tasks \cite{zeng2022transformers, das2023long}.
% We keep a shortcut for the linear part in each block. When all residual connections are activated (e.g., when graph relation is unnecessary), our model can degenerate into a family of channel-independent linear models, e.g., \cite{zeng2022transformers,das2023long}, which show great potential for long-term series forecasting.

\paragraph{{Parameter Sharing}}
In contrast to recent design trends that use separate node parameters or graphs for different layers \cite{zhang2020spatio,oreshkin2021fc}, we use a globally shared node embedding for all modules, including the \textsc{TimeMixer} and \textsc{SpaceMixer}.
This aims to simplify the end-to-end training of random embedding while reducing the model size. Furthermore, \cite{yang2022graph} and \cite{han2022mlpinit} showed that MLP and GNN can share similar feature spaces. Without the contextualization function, \textsc{SpaceMixer} can collapse into \textsc{TimeMixer}. Considering this, we instantiate consecutive \textsc{SpaceMixer} layers with a shared feedforward weight. 
This parameter-sharing design is inspired by the connection between MLP and GNN in spatiotemporal graphs.
% Different from recent design trends that assign separate node parameters or graphs for different layers \cite{zhang2020spatio,oreshkin2021fc}, we set the NE globally shared for all modules. 
% The goal is to facilitate end-to-end training of random node features while reducing the model size. Furthermore, recent studies have empirically shown that MLPs and MPNNs can share similar feature spaces \cite{yang2022graph,han2022mlpinit}. In the absence of MP, \textsc{SpaceMixer} can collapse into \textsc{TimeMixer}. Taking this into account, we instantiate successive MP layers with a shared linear transform for time mixing. 

\paragraph{{Shallow-Layer Structure}}
Instead of using deep multilayer architectures, we use shallow layer structures with larger receptive fields. Specifically, we adopt a small number of structured space mixing layers (e.g., 1 layer in most cases) with all-to-all connections to capture long-range interactions. This approach avoids the use of multiple stacking of sparse graph aggregators or hierarchical operations such as diffusion convolutions \cite{li2018diffusion}.
% This treatment is consistent with the previous studies' finding that the expressive power of GNNs is determined by both depth and width \cite{loukas2019graph}, and wider GNNs can be more expressive than deeper ones \cite{wu2019simplifying}. In other words, when the graph is dense, a single global MP can gather adequate information adaptively from arbitrary nodes. Therefore, despite its simple structure, our model has sufficient expressivity to capture pairwise interactions.
This treatment is consistent with the previous finding that wider graphs can be more expressive than deeper ones under some conditions \cite{wu2023simplifying}. In other words, when the graph is dense, a single global mixing can gather adequate information adaptively from arbitrary nodes. Therefore, despite the simple structure, it has sufficient expressivity and a large receptive field to capture pairwise interactions. 
% Another crucial aspect to consider when utilizing a dense relation graph is that it prompts the model to gather comprehensive information from other series in order to obtain a precise ``when locator.'' 
Another crucial aspect to consider when utilizing an all-to-all connection is that it prompts the model to gather comprehensive information from other windows to obtain a precise ``temporal context.''

\begin{table}[!htbp]
\centering
% \begin{minipage}[c]{0.49\textwidth}
\centering
\caption{Statistics of public urban computing benchmarks.}
\label{tab_datasets}
\small
\centering
\begin{small}
\setlength{\tabcolsep}{1pt}
\resizebox{1\columnwidth}{!}{
\begin{tabular}{l|c| c| c| c| c| c}
\toprule
 \multicolumn{2}{c|}{\textsc{Datasets}} & \textsc{Type} & \textsc{Steps} & \textsc{Nodes} & \textsc{Edges} & \textsc{Interval}\\
\midrule
\multirow{8}{*}{\rotatebox{90}{Traffic}} &
\texttt{METR-LA} &  speed & 34,272 & 207 & 1,515 & 5 min \\
& \texttt{PEMS-BAY} &  speed & 52,128 & 325 & 2,369 & 5 min \\
% \midrule
& \texttt{PEMS03} &  volume & 26,208 & 358 & 546 & 5 min \\
& \texttt{PEMS04} &  volume & 16,992 & 307 & 340 & 5 min \\
& \texttt{PEMS07} &  volume & 28,224 & 883 & 866 & 5 min \\
& \texttt{PEMS08} &  volume & 17,856 & 170 & 277 & 5 min \\
% \midrule
& \texttt{TrafficL} & occupancy & 17,544 & 862 & - & 60 min \\
% \midrule
& \texttt{LargeST-GLA} & volume & 525,888 & 3,834 & 98,703 & 15 min \\
% \midrule
% \multirow{3}{*}{\rotatebox{90}{Energy}} &
% \texttt{Electricity} & electricity & 26,304 & 321 & - & 60 min \\
% & \texttt{CER-EN} &  smart meters & 52,560 & 6,435 & 639,369 & 10 min \\
% & \texttt{PV-US} &  solar power & 52,560 & 5,016 & 417,199 & 10 min \\
% \midrule
% \multirow{3}{*}{\rotatebox{90}{Environ.}} &
% \texttt{AQI} &  pollutant & 8,760 & 437 & 2,699 & 60 min \\
% & \texttt{Global Temp} & temperature & 17,544 & 3,850 & - & 60 min \\
% & \texttt{Global Wind} &  wind speed & 17,544 & 3,850 & - & 60 min \\
\bottomrule
\end{tabular}}
\end{small}
% \end{minipage}
\end{table}

\begin{table*}[t]
\caption{Full results on \texttt{METR-LA, PEMS-BAY, PEMS03, PEMS04, PEMS07, and PEMS08} datasets. MAE for \{15, 30, 60\} minutes forecasting horizons, as well as MAE, MSE, and MAPE averaged over one hour (12 time steps) are reported.}
\begin{center}
\small
\setlength{\tabcolsep}{3pt}
\resizebox{1\textwidth}{!}{
\begin{tabular}{c| c | ccc | cc | ccc | cc | ccc | cc }
\cmidrule[1.5pt]{1-17}
 \multicolumn{2}{c}{\textsc{Dataset}} & \multicolumn{5}{c|}{\texttt{METR-LA}} & \multicolumn{5}{c|}{\texttt{PEMS-BAY}} & \multicolumn{5}{c}{\texttt{PEMS03}} \\
\cmidrule[0.8pt]{1-17}
& \multicolumn{1}{c|}{\multirow{2}{*}[-0.45em]{\textsc{Metric}}}
& 15 min & 30 min & 60 min & \multicolumn{2}{c|}{Average} & 15 min & \multicolumn{1}{c}{30 min} & \multicolumn{1}{c|}{60 min} & \multicolumn{2}{c|}{Average} & 15 min & 30 min & 60 min & \multicolumn{2}{c}{Average}\\
\cmidrule[0.5pt]{3-17}
& &  MAE &  MAE &  MAE &  MAE &  MAPE (\%) &   MAE &  MAE &  MAE &  MAE  &  MAPE (\%) &  MAE &  MAE &  MAE &  MAE  &  MAPE (\%) \\
\midrule
\multirow{12}{*}{\rotatebox{90}{STGNNs \& Transformers}} & {AGCRN} & 2.85  & 3.19  & 3.56  &  3.14  & 8.69  & 1.38  & 1.69  & 1.94  & 1.63    & 3.71 & 14.65  & 15.72  & 16.82  &  15.58  & 15.08 \\
&  {DCRNN} & 2.81  & 3.23  & 3.75  & 3.20   & 8.90  & 1.37  & 1.72  & 2.09  & 1.67    & 3.78   & 14.59  & 15.76  & 18.18  & 15.90  & 15.74  \\
& {GWNet} & 2.74  & 3.14  & 3.58  & 3.09   & 8.52  & 1.31  & 1.65  & 1.96  & 1.59   & 3.57 & 13.67  & 14.60  & 16.23  & 14.66  & 14.88  \\
& {GatedGN} & 2.72  & 3.05  & 3.43  & 3.01   & 8.20  & 1.34  & 1.65  & 1.92  & 1.58   & 3.55 & 13.72  & 15.49  & 19.08  & 17.08  & 14.44\\
& {GRUGCN} & 2.93  & 3.47  & 4.24  & 3.46  & 9.85  & 1.39  & 1.81  & 2.30  & 1.77  & 4.03   & 14.63  & 16.44  & 19.87  & 16.62  & 15.96\\
& {EvolveGCN} & 3.27  & 3.82  & 4.59  & 3.81  & 10.56  & 1.53  & 2.00  & 2.53  & 1.95  & 4.43 & 17.07  & 19.03  & 22.22  & 19.11  & 18.61\\
 &{ST-Transformer} & 2.97  & 3.53  & 4.34  & 3.52  & 10.06  & 1.39  & 1.81  & 2.29  & 1.77   & 4.09  & 14.03  & 15.72  & 18.74  & 15.85  & 15.37  \\
& {STGCN} & 2.84  & 3.25  & 3.80  & 3.23  & 9.02  & 1.37  & 1.71  & 2.08  & 1.66  & 3.75 & 14.27  & 15.49  & 18.02  & 15.61  & 16.07\\
& {MTGNN} & 2.79  & 3.12  & 3.46  &  3.07  & 8.57  & 1.34  & 1.65  & 1.91  & 1.58   & 3.51 & 13.74  & 14.76  & 16.13  &  14.70  &  14.90 \\
& {D2STGNN} & 2.74  &  3.08  & 3.47  &  3.05  & 8.44  & \textbf{1.30}  & \textbf{1.60}  & 1.89  & 1.55   & 3.50 & 13.36  &  14.45  & 15.96  &  14.45  &  14.55 \\
 &{DGCRN} & 2.73  &  3.10  & 3.54  &  3.07  & 8.39  & 1.33  & 1.65  & 1.95  & 1.59  & 3.61 & 13.83  &  14.77  & 16.14  &  14.73  &  14.86 \\
& {SCINet} & 3.06  & 3.47  & 4.09  &  3.47 & 9.93  & 1.52  & 1.85  & 2.24  & 1.82  & 4.14 & 14.39  & 15.25 & 17.37  &  15.43  &  15.37 \\
\midrule
\multirow{4}{*}{\rotatebox{90}{MLPs}}&{FreTS} & 3.03  & 3.67  & 4.62 & 3.67  & 10.49 & 1.41 & 1.87 &2.45 & 1.84  & 4.23 & 14.53  & 16.78 & 21.07  & 17.04 & 16.18  \\
& {STID} & 2.82  & 3.18  & 3.56  & 3.13  & 9.07  & 1.32  & 1.64  & 1.93  & 1.58  & 3.58 & 13.88  & 15.26  & 17.41  & 15.27  & 16.39  \\
&{TSMixer} & 2.93  &  3.31 & 3.79 &  3.28  & 8.97 & 1.44 & 1.80 & 2.12 &  1.73  &  3.91& 14.47  & 15.34 & 17.11  & 15.43 & 15.41  \\
% \midrule
\cmidrule[0.8pt]{2-17}
 & \textbf{\gls{model}} & \textbf{2.66} & \textbf{2.99 } & \textbf{3.36 } & \textbf{2.95 }  & \textbf{8.04} & \textbf{1.30 } & \textbf{1.60 } & \textbf{1.86 } & \textbf{1.54}  & \textbf{3.45} & \textbf{13.15} & \textbf{14.20} & \textbf{15.79} & \textbf{14.18} & \textbf{13.77} \\
\bottomrule
\end{tabular}}

\resizebox{1\textwidth}{!}{
\begin{tabular}{c|c | ccc | cc | ccc | cc | ccc | cc }
\cmidrule[1.5pt]{1-17}
 \multicolumn{2}{c}{\textsc{Dataset}} & \multicolumn{5}{c|}{\texttt{PEMS04}} & \multicolumn{5}{c|}{\texttt{PEMS07}} & \multicolumn{5}{c}{\texttt{PEMS08}} \\
\cmidrule[0.8pt]{1-17}
& \multicolumn{1}{c|}{\multirow{2}{*}[-0.45em]{\textsc{Metric}}}
& 15 min & 30 min & 60 min & \multicolumn{2}{c|}{Average} & 15 min & \multicolumn{1}{c}{30 min} & \multicolumn{1}{c|}{60 min} & \multicolumn{2}{c|}{Average} & 15 min & 30 min & 60 min & \multicolumn{2}{c}{Average}\\
\cmidrule[0.5pt]{3-17}
& &  MAE &  MAE &  MAE &  MAE  &  MAPE (\%) &   MAE &  MAE &  MAE &  MAE  &  MAPE (\%) &  MAE &  MAE &  MAE &  MAE  &  MAPE (\%) \\
\midrule
\multirow{12}{*}{\rotatebox{90}{STGNNs \& Transformers}} & {AGCRN} &18.27  & 18.95  & 19.83  & 18.90  & 12.69 & 19.55  & 20.65  & 22.40  &  20.64  & 9.42  & 14.50  & 15.16  & 16.41  & 15.23  & 10.46  \\
& {DCRNN} &19.19  & 20.54  & 23.55  & 20.75  & 14.32 & 20.47  & 22.11  & 25.77  & 22.30  & 9.51  & 14.84  & 15.93  & 18.22  & 16.06  & 10.40  \\
 &{GWNet}&18.17  & 18.96  & 20.23  & 18.95  & 13.59 & 19.87  & 20.97  & 23.53  & 21.13  & 9.96  & 14.21  & 14.99  & 16.45  & 15.02  & 9.70  \\
& {GatedGN} &18.10  & 18.84  & 20.03  & 18.81  & 13.11 & 21.01  & 22.74  & 25.81  & 22.68  & 10.18  & 14.07  & 14.96  & 16.27  & 14.91  & 9.64  \\
& {GRUGCN} &19.89  & 22.28  & 27.37  & 22.68  & 15.81  & 21.15  & 24.00  & 29.91  & 24.37  & 10.29  & 15.46  & 17.32  & 21.16  & 17.55  & 11.43  \\
& {EvolveGCN} &23.21  & 25.82  & 30.97  & 26.21  & 17.79 & 24.72  & 28.09  & 34.41  & 28.40  & 12.11  & 18.21  & 20.46  & 24.45  & 20.64  & 13.11  \\
& {ST-Transformer}& 19.13  & 21.33  & 25.69  & 21.63  & 15.12 & 20.17  & 22.80  & 27.74  & 23.05  & 9.77 & 14.39  & 15.87  & 18.69  & 16.00  & 10.71  \\
& {STGCN}& 19.26  & 20.75  & 24.03  & 20.95  & 14.79  & 20.26  & 22.33  & 26.58  & 22.53  & 9.66  & 14.81  & 16.02  & 18.76  & 16.20  & 10.57  \\
& {MTGNN}& 17.88  & 18.49  & 19.62  & 18.48  & 12.76 & 18.94  & 20.23  & 22.23  &  20.19  & 8.59 & 13.80  & 14.57  & 15.82  & 14.57  & 9.46   \\
& {D2STGNN}& 17.60  & 18.39  & 19.63  & 18.39  & 12.65 & 18.49  & 20.74  & 23.10  &  20.84  &  9.99  & 13.65  & 14.53  & 16.00  & 14.52  & 9.40   \\
& {DGCRN}& 18.05  & 18.76  & 20.07  & 18.77  & 13.13 & 18.58  &  21.10  & 22.55  &  21.24  &  10.45  & 13.96  & 14.68  & 15.96  & 14.70  & 9.62 \\
& {SCINet}& 18.30  & 19.03  & 20.85  & 19.19  & 13.31 & 21.63  &  22.89  & 25.96 &  23.11  &  10.00  & 14.57  & 15.60  & 17.75  & 15.77  & 10.13 \\
\midrule
\multirow{4}{*}{\rotatebox{90}{MLPs}}
& {FreTS} & 19.91  &  22.64 &28.23  &  23.06 &  15.74 & 21.21 & 24.66 & 31.45 & 25.05 &  10.82 & 15.37  & 17.62 & 22.11  &17.92  &   11.75   \\
& {STID}& 17.77  & 18.60  & 20.01  &  18.60  & 12.87 & 18.66  & 19.93  & 21.91  & 19.88  & 8.82  & 13.53  & 14.29  & 15.75  & 14.32  & 9.69   \\
& {TSMixer} & 19.12  & 19.78  & 21.03 & 19.81  & 13.40  & 20.15 & 21.73 & 24.67 &21.81 & 8.98  &  15.84 &  16.82& 18.80  & 17.00 &  12.37   \\
% \midrule
\cmidrule[0.8pt]{2-17}
 &  \textbf{\gls{model}} &\textbf{17.37} & \textbf{18.05} & \textbf{19.14} & \textbf{18.03} & \textbf{12.34} & \textbf{18.15} & \textbf{19.34} & \textbf{21.05} & \textbf{19.28} & \textbf{8.05} & \textbf{13.17} & \textbf{13.96} & \textbf{15.28} & \textbf{13.98} & \textbf{9.02} \\
\bottomrule
\end{tabular}}
\begin{tablenotes} 
% \tiny
\scriptsize
{
\item - Best results are bold marked. Note that - indicates the model runs out of memory with the minimum batch size. 
}
\end{tablenotes} 
% \vspace{-0.2cm}
\label{tab:traffic_results_all}
\end{center}
\end{table*}

\section{Evaluation on Public Benchmarks}\label{sec:exp}
Extensive experiments are carried out to compare \gls{model} with state-of-the-art neural forecasting baselines in 8 well-known traffic forecasting benchmarks. These benchmarks included short-term, long-term, and large-scale forecasting tasks.
Brief statistics information is provided in Tab. \ref{tab_datasets}, and detailed descriptions can be found in the Appendix. To demonstrate the methodological generality, we also perform supplementary experiments in the Appendix using data from other sources, including energy record and environmental measurements.
To validate its effectiveness in real-world applications, we further conduct a collaborative urban congestion project with Baidu in 5 large-scale urban road networks.
Experimental setups and full results of \textbf{traffic data} are summarized below. Details and supplementary results on other data are given in the Appendix.
% For a comprehensive performance evaluation, we compare our \gls{model} model with 15 neural forecasting baselines on seven well-known traffic forecasting benchmark tasks. The experimental setups and results are summarized below. Further details and supplementary results can be found in the appendix.
% Extensive experiments are carried out to assess the performance of the \gls{model} model on 7 well-known traffic forecasting benchmarks and compare it with 15 neural forecasting baselines. Details regarding the experimental settings can be found in the Appendix.
{
Model implementations using PyTorch are publicly available at: \textbf{\url{https://github.com/tongnie/NexuSQN}}. }

\subsection{Experiment Setup}
% \paragraph{Datasets} 

\paragraph{Traffic Flow Benchmarks} 
Numerous experiments were conducted to comprehensively evaluate models' predictive capabilities. 
(1) To evaluate the short-term forecasting performance, we adopt six high-resolution highway traffic flow datasets, including two traffic speed datasets: \texttt{METR-LA} and \texttt{PEMS-BAY} ~\cite{li2018diffusion}, and four traffic volume datasets: \texttt{PEMS03}, \texttt{PEMS04}, \texttt{PEMS07}, and \texttt{PEMS08}~\cite{guo2021learning}. 
The six datasets contain traffic measurements aggregated every 5 minutes from loop sensors installed on highway networks. \texttt{METR-LA} contains spot speed data from 207 loop sensors over a period of 4 months from Mar 2012 to Jun 2012, located at the Los Angeles County highway network. \texttt{PEMS-BAY} records 6 months of speed data from $325$ static detectors in the San Francisco South Bay Area. \texttt{PEMS0X} contains the real-time highway traffic volume information in California, collected by the Caltrans Performance Measurement System (PeMS) \cite{chen2001freeway} in every 30 seconds. The raw traffic flow is aggregated into a 5-minute interval for the experiments. Similarly to \texttt{METR-LA} and \texttt{PEMS-BAY}, \texttt{PEMS0X} also includes an adjacency graph calculated by the physical distance between the sensors.
Additionally, adjacency matrices constructed from the geographical distance between the sensors are prepared for baselines that require predefined graphs \cite{wu2019graph}.
(2) To evaluate NexuSQN's long-term forecasting performance, 
we use the \texttt{TrafficL} benchmark \cite{lai2018modeling}. This dataset provides hourly records of road occupancy rates (between 0 and 1) measured by 862 sensors on San Francisco Bay Area freeways over a period of 48 months. 
% \paragraph{Traffic Occupancy Datasets} \texttt{TrafficL} is a common benchmark for the long-term series forecasting (LTSF) task. 
(3) To verify the scalability of models, we adopt the large-scale \texttt{LargeST} benchmarks. \texttt{LargeST} dataset contains 5 years of traffic readings from 01/01/2017 to 12/31/2021 collected every 5 minutes by 8,600 traffic sensors in California. We adopt the largest subset \texttt{GLA} and use the readings from 2019 that are considered, aggregated into 15-minute intervals.

\paragraph{Baselines} We consider a variety of baseline models from the literature, including STGNNs for shot-term forecasting:
\texttt{STGCN} \cite{yu2017spatio}; 
\texttt{DCRNN} \cite{li2018diffusion}; 
\texttt{GWNet} \cite{wu2019graph}; 
\texttt{AGCRN} \cite{bai2020adaptive};
% \texttt{EvolveGCN} \cite{pareja2020evolvegcn}; 
\texttt{ST-Transformer} \cite{xu2020spatial}; 
\texttt{MTGNN} \cite{wu2020connecting}; 
\texttt{GatedGN} \cite{satorras2022multivariate}; 
\texttt{GRUGCN} \cite{gao2022equivalence}; 
\texttt{STID} \cite{shao2022spatial}; 
\texttt{D2STGNN} \cite{shao2022decoupled}; 
\texttt{DGCRN} \cite{li2023dynamic}. And long-term series forecasting models:
\texttt{Autoformer} \cite{wu2021autoformer}; 
\texttt{Informer} \cite{zhou2021informer}; 
\texttt{FEDformer} \cite{zhou2022fedformer}; 
\texttt{Pyraformer} \cite{liu2022pyraformer};
\texttt{StemGNN} \cite{cao2020spectral};
\texttt{SCINet} \cite{liu2022scinet};
\texttt{DLinear} \cite{zeng2022transformers}; 
\texttt{PatchTST} \cite{nie2022time}; 
\texttt{TimesNet} \cite{wu2022timesnet}, etc.

\begin{table}[!htbp]
\centering
\caption{Input and output settings.}
\label{tab_settings}
\small
\centering
\begin{small}
\setlength{\tabcolsep}{5pt}
\resizebox{0.8\columnwidth}{!}{
\begin{tabular}{l|c| c| c| c}
\toprule
 \multicolumn{2}{c|}{\textsc{Datasets}} & \textsc{window} & \textsc{Horizon} & \textsc{Graphs} \\
\midrule
\multirow{8}{*}{\rotatebox{90}{Traffic}} &
\texttt{METR-LA} &  12 & 12 & True  \\
& \texttt{PEMS-BAY} &  12 & 12 & True \\
& \texttt{PEMS03} &  12 & 12 & True  \\
& \texttt{PEMS04} &  12 & 12 & True  \\
& \texttt{PEMS07} &  12 & 12 & True  \\
& \texttt{PEMS08} &  12 & 12 & True  \\
& \texttt{TrafficL} & 96 & 384 & False \\
& \texttt{LargeST-GLA} & 12 & 12 & True  \\
% \midrule
% \multirow{3}{*}{\rotatebox{90}{Energy}} &
% \texttt{Electricity} & 336 & 96 & False \\
% & \texttt{CER-EN} &  36 & 22 & True  \\
% & \texttt{PV-US} &  36 & 22 & True \\
% \midrule
% \multirow{3}{*}{\rotatebox{90}{Environ.}} &
% \texttt{AQI} &  24 & 3 & True \\
% & \texttt{Global Temp} & 48 & 24 & False  \\
% & \texttt{Global Wind} &  48 & 24 & False  \\
\bottomrule
\end{tabular}}
\end{small}
\end{table}

\paragraph{Basic Setups} We adopt the widely used input-output settings in related literature to evaluate the models. These settings are given in Tab. \ref{tab_settings}.
For all the datasets, we adopt the same training (70\%), validating (10\%), and testing (20\%) set splits and pre-processing steps as in previous work. 
It is worth commenting that \gls{model} does not rely on redefined graphs, and results in Tab. \ref{tab_ablation} indicate that the benefits of incorporating predefined graphs are marginal.
Time-of-day information is provided as exogenous variables for our model, and day-of-week feature is input to baselines that require this information. In addition, labels with nonzero values are used to compute the metrics. And the performance of all methods is recorded in the same evaluation environment. We evaluate the model's performance using metrics such as mean absolute error (MAE), mean squared error (MSE), and mean absolute percentage error (MAPE). For long-term benchmarks including \texttt{TrafficL} and \texttt{Electricity}, we do not use the standardization on labels and use the normalized mean absolute error (NMAE), mean relative error (MRE), and mean absolute percentage error (MAPE) as metrics.

\paragraph{Platform} All these experiments are conducted on a Windows platform with one single NVIDIA RTX A6000 GPU with 48GB memory. Our implementations and suggested hyperparameters are mainly based on PyTorch \cite{paszke2019pytorch}, Torch Spatiotemporal \cite{Cini_Torch_Spatiotemporal_2022}, and BasicTS \cite{shao2023exploring} benchmark tools.

\paragraph{Hyperparameters} {To ensure a fair and unbiased comparison, we adopt the original implementations presented in each paper \cite{li2018diffusion,wu2019graph,bai2020adaptive,satorras2022multivariate,pareja2020evolvegcn,yu2017spatio,wu2020connecting,shao2022decoupled,li2023dynamic,liu2022scinet,yi2023frequency,shao2022spatial}}. {We also use the recommended hyperparameters for the baselines from the public benchmark tools such as BasicTS \cite{liang2022basicts} and TorchSpatiotemporal \cite{Cini_Torch_Spatiotemporal_2022}.} During the evaluation, all baselines are trained, validated, and tested in identical environments.
Our \gls{model} contains only a few model hyperparameters and is easy to tune. 
% The detailed configurations of traffic datasets are shown in Tab. \ref{tab:hyp}. 
And the hyperparameters of other datasets can be found in our repository.
The configurations of other baseline models and implementations follow the official resources as much as possible.

\subsection{Results on Traffic Benchmarks}
\paragraph{Short-term Performance}
Results of model comparison on short-term traffic flow benchmarks are shown in Table \ref{tab:traffic_results_all}. In general, our model achieves comparable or even superior performance compared to its more complicated counterparts. 
% Notably, forecasting the \texttt{METR-LA} data is considered challenging due to its complex temporal patterns. However, our \gls{model} consistently outperforms baselines by a large margin, even without popular sequential techniques. For the \texttt{PEMS-BAY} data, \gls{model} remains among the best-performing predictors. 
% Compared to \texttt{STID}, \gls{model} features the extraction of distinguishable low-frequency node representations with more effective ``when locators'', thus showing better accuracy in all scenarios. 
Compared to other MLP-based models, \gls{model} features the exploitation of relational structures with distinguishable space mixing, thus showing better accuracy in all scenarios. 
Additionally, advanced STGNNs such as \texttt{MTGNN}, and \texttt{D2STGNN} also show competitive results. 
% However, high memory consumption and computational complexity hinder them from large-scale applications.
However, high computational complexity requires them to consume much more memory and training times.
% Similar observations can be made in the traffic volume datasets. 
Although our model does not depend on predefined graphs, it still achieves promising performance in traffic data. This observation echoes our previous finding that although distance-based adjacency matrices may reflect traffic speed patterns, they are not sufficient to understand traffic volume patterns \cite{nie2023towards}. 
Finally, it is important to note that our model only contains \textbf{one single} \textsc{SpaceMixer} layer in all traffic benchmarks. This finding surprisingly demonstrates the effectiveness of a parsimonious structure for traffic forecasting.

\begin{table}[!htbp]
\centering
\caption{Long-term results on \texttt{TrafficL} (in: 96, out: 384).}\label{tab_traffic_occ}
  \centering
  \begin{small}
  \setlength{\tabcolsep}{5pt}
  \renewcommand{\multirowsetup}{\centering}
  \resizebox{0.8\columnwidth}{!}{

\begin{tabular}{l | cccc | c cc}
% \cmidrule[1pt]{2-6}
\toprule
 \multicolumn{1}{c}{\textsc{Dataset}} & \multicolumn{6}{c}{\texttt{TrafficL}}  \\
% \cmidrule[0.8pt]{2-6}
\midrule
\multicolumn{1}{c|}{\multirow{2}{*}[-0.45em]{\textsc{Metric}}} & 96 & 192 & 288 & 384 & \multicolumn{2}{c}{Ave.}  \\
\cmidrule[0.5pt]{2-7}
% \midrule
   & \scriptsize MAE & \scriptsize MAE & \scriptsize MAE & \scriptsize MAE & \scriptsize NMAE & \scriptsize MRE \\
 \midrule
{Autoformer} & 1.34 &1.43 &1.82 & 2.07 & 2.38 & 27.90 \\
{Informer} & 1.35 & 1.44 & 1.80& 2.00 & 2.35 & 27.72\\
{TimesNet}  & 1.31  & 1.41  & 1.43  & 1.52  & 1.88  & 22.60  \\
{SCINet}  & 1.21  & 1.40  & 1.41  &  1.47  & 1.86 & 22.27 \\
\midrule
{DLinear} & 1.30  & 1.41  & 1.74   &   1.92  & 2.14 & 25.73   \\
\textbf{\gls{model}}  & \textbf{1.11}  &  \textbf{1.18}  &  \textbf{1.19}   &   \textbf{1.28}  & \textbf{1.63}  & \textbf{19.55} \\
\bottomrule
\end{tabular}
  }
  \end{small}
\begin{tablenotes} 
% \tiny
\scriptsize
{
\item - We do not use normalization on the raw data and report the original metrics.
}
\end{tablenotes} 
\end{table}

\begin{table}[!htbp]
\caption{Large-scale results on \texttt{LargeST} (in: 12, out: 12).}
\small
\begin{center}
\setlength{\tabcolsep}{5pt}
\resizebox{0.8\columnwidth}{!}{
\begin{tabular}{l | ccc | c c}
% \cmidrule[1pt]{2-6}
\toprule
 \multicolumn{1}{c}{\textsc{Dataset}} & \multicolumn{5}{c}{\texttt{LargeST-GLA}}  \\
% \cmidrule[0.5pt]{2-6}
\midrule
 \multicolumn{1}{c|}{\multirow{2}{*}[-0.45em]{\textsc{Metric}}} & 3 & 6 & 12 & \multicolumn{2}{c}{Ave.}  \\
\cmidrule[0.5pt]{2-6}
 & \scriptsize MAE & \scriptsize MAE & \scriptsize MAE & \scriptsize MAE & \scriptsize Time (h) \\
 \midrule
{AGCRN} & 18.64  &  22.01  & 26.91   &   21.99  &  38  \\
{DCRNN}  & 18.64  & 23.58  & 31.41  & 23.68  & 69  \\
{GWNet}  & 16.42  & 19.87  & 24.59  &  19.70  & 31  \\
% {GRUGCN}  & -  &  -  & -   &  - &  -   \\
% {EvolveGCN}  & -  &  -  & -   &  -  &  -  \\
% {TSMixer}  & 19.16  &  24.69  &  33.33   &  24.76 &  3.5   \\
% {STID}  & 17.07  &  20.64  & 25.46   &  20.47  &  3  \\
{MTGNN}  & 17.30  &  20.24  & 24.56   &  20.18 &  26   \\
% {GatedGN}  & -  &  -  & -   &  -  &  -  \\
\midrule
{TSMixer}  & 19.16  &  24.69  &  33.33   &  24.76 &  3.5   \\
\textbf{\gls{model}}  & \textbf{16.15}  &  \textbf{19.48}  & \textbf{24.09}   &  \textbf{19.33} & \textbf{1.5}  \\
\bottomrule
\end{tabular}}
\label{tab:largest_result}
\end{center}
\end{table}

\paragraph{Long-term Performance}
We also evaluate the model's performance in long-term forecasting. Table \ref{tab_traffic_occ} shows the results for different prediction horizons with a historical series of 96 steps. Compared to three strong baselines in the LTSF literature, \gls{model} shows great potential in this challenging task. This can be attributed to the encoding of both relative timestamps and local patterns.

% We also assess the ability to predict long-term series. Tab. \ref{tab_traffic_occ} reports the results of different prediction horizons with a 96-step sequence as input. Compared to three strong baselines in the LTSF literature, \gls{model} shows great potential to handle the challenging LTSF task. This may be attributed to the model’s encoding of both relative time stamps and local patterns.
% because our model can encode series with both relative time stamps and local patterns.

\paragraph{Large-Scale Forecasting Performance} We adopt the \texttt{LargeST} data \cite{liu2023largest} to evaluate the performance on large-scale traffic networks. Tab. \ref{tab:largest_result} shows that even with a single \textsc{SpaceMixer} layer, our model can outperform the complicate STGNNs. Furthermore, many advanced models such as \texttt{D2STGNN} and \texttt{DGCRN} are not applicable due to high memory consumption. Instead, \gls{model} can converge in less than 1.5 hours on such a billion-level dataset, which is an \textbf{order of magnitude speed-up ($\geq$10x)} compared to the STGNNs.

% \begin{table}[!htbp]
% \caption{Results on \texttt{LargeST} benchmarks. Note that - indicates the model runs out of memory with the minimum batch size. Best results are bold marked. \textbf{\textcolor{purple}{[TO BE DONE]}}}
% \small
% \begin{center}
% \setlength{\tabcolsep}{3pt}
% \scalebox{0.9}{
% \begin{tabular}{l | ccc | c c}
% \cmidrule[1pt]{2-6}
%  \multicolumn{1}{c}{} & \multicolumn{5}{c}{\texttt{LargeST-GLA}}  \\
% \cmidrule[0.8pt]{2-6}
%  \multicolumn{1}{c}{} & 45 min & 90 min & 180 min & \multicolumn{2}{c}{Ave.}  \\
% \cmidrule[0.5pt]{2-6}
%    \multicolumn{1}{c}{} & \scriptsize MAE & \scriptsize MAE & \scriptsize MAE & \scriptsize MAE & \scriptsize MAPE \\
%  \midrule
% {AGCRN} & 18.64  &  22.01  & 26.91   &   21.99  &  16.09  \\
% {DCRNN}  & 18.64  & 23.58  & 31.41  & 23.68  & 15.60  \\
% {GWNet}  & 16.42  & 19.87  & 24.59  &  19.70  & 12.48  \\
% % {GRUGCN}  & -  &  -  & -   &  - &  -   \\
% % {EvolveGCN}  & -  &  -  & -   &  -  &  -  \\
% {STGCN}  & 19.16  &  24.69  &  33.33   &  24.76 &  17.63   \\
% {STID}  & 17.07  &  20.64  & 25.46   &  20.47  &  14.54  \\
% {MTGNN}  & 17.30  &  20.24  & 24.56   &  20.18 &  13.29   \\
% % {GatedGN}  & -  &  -  & -   &  -  &  -  \\
% \midrule
% \textbf{\gls{model}}  & \textbf{16.15}  &  \textbf{19.48}  & 24.09   &  19.33 &12.34  \\
% \bottomrule
% \end{tabular}}
% \label{tab:largest_result}
% \end{center}
% \end{table}

\begin{figure*}[!htbp]
    \centering
    \includegraphics[width=0.9\textwidth]{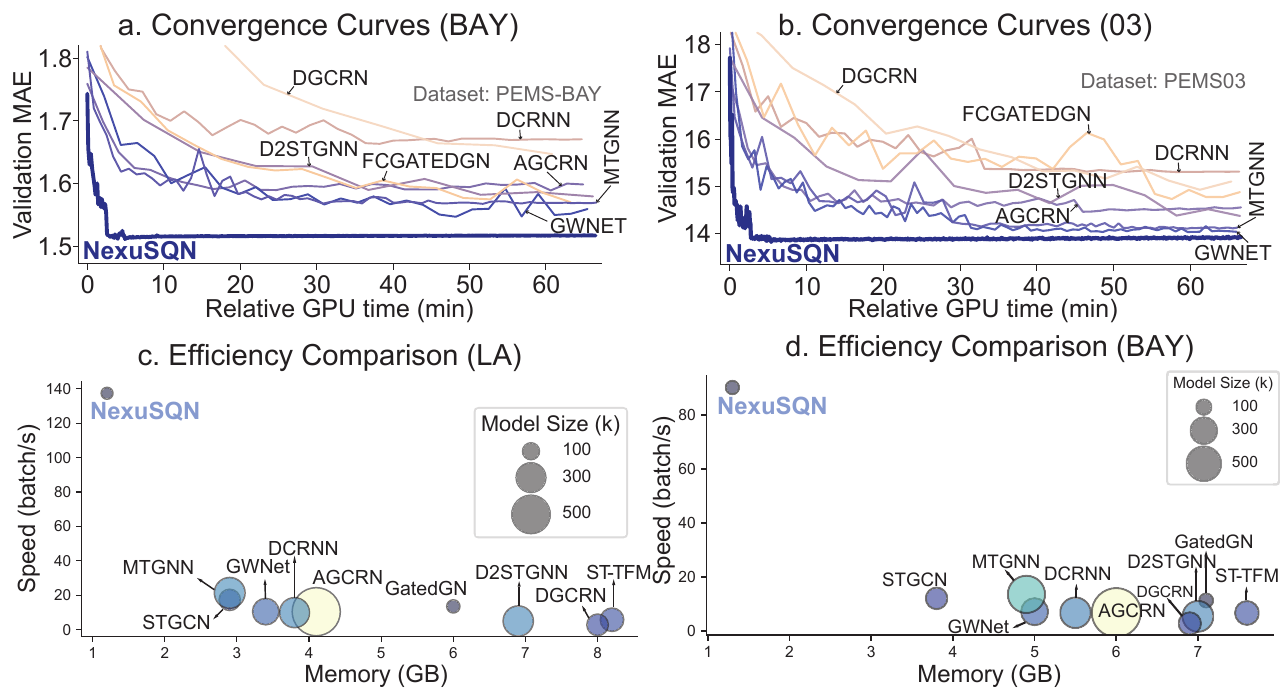}
    \caption{Computational performance.}
    \label{fig_sup:efficiency_all}
\end{figure*}

\begin{table}[!htbp]
\caption{Model computational performances.} 
\begin{center}
\setlength{\tabcolsep}{3pt}
\resizebox{0.9\columnwidth}{!}{
\begin{tabular}{l | cccc | cccc }
\cmidrule[1pt]{1-9}
\multicolumn{1}{c}{\textsc{Datasets}} & \multicolumn{4}{c|}{\texttt{METR-LA}} & \multicolumn{4}{c}{\texttt{PEMS-BAY}}  \\
\cmidrule[0.8pt]{1-9}
   \multicolumn{1}{c}{\textsc{Metrics}} &  \makecell{Speed \\ (Batch/s)} &  \makecell{Memory \\ (GB)}&  \makecell{Batch \\ size} &  \makecell{Model\\ size (k)}&   \makecell{Speed \\ (Batch/s)} &  \makecell{Memory \\ (GB)}&  \makecell{Batch \\ size} &  \makecell{Model\\ size (k)}\\
 \midrule
{AGCRN} & 10.43 & 4.1  & 64 & 989  & 6.78 & 6.0   & 64 & 991  \\
 {DCRNN} & 9.90  & 3.8   & 64 & 387  & 6.72 & 5.5   & 64  &387   \\
{GWNet} & 10.50 & 3.4  & 64 & 301  & 7.11 & 5.0  & 64 &  303  \\
{GatedGN} & 13.53 & 6.0  & 32 & 74.6  & 11.27 & 7.1  & 16 & 82.2   \\
% {\footnotesize GRUGCN} & 97.09  & 2.5  & 64 & 234  & 56.51 & 2.9  & 64 &234   \\
% {\footnotesize EvolveGCN} & 49.66 & 1.8  & 64 & 208 & 38.36 & 2.2  & 64 &208   \\
{ST-Transformer} & 5.77 & 8.2  & 64 & 236  & 6.63 & 7.6  & 32 & 236  \\
{STGCN} & 17.31 & 2.9  & 64 & 194  & 12.05 & 3.8  & 64 & 194  \\
% {\footnotesize STID} & 98.18 & 1.6  & 64 & 208  & 77.81 & 1.7  & 64 & 230   \\
{MTGNN} & 21.39 & 2.9  & 64 & 405  &  13.54 &  4.9  & 64 & 573   \\
{D2STGNN} & 5.02 &  6.9  & 32 & 392  &  5.38 &  7.0  & 16  &394  \\
{DGCRN} & 2.82 &  8.0  & 64 &199  &  2.74 &  6.9  & 32 &208   \\
% {\footnotesize SCINet} & 76.72 &  2.6  & 64 &121  &  78.12 &  2.7  & 64 & 316   \\
\midrule
 \textbf{\gls{model}} & \textbf{137.35} & \textbf{1.2} & \textbf{64} & \textbf{60.5} & \textbf{90.26} & \textbf{1.3} & \textbf{64} & \textbf{75.6} \\
\bottomrule
\end{tabular}}
\label{tab_sup:gpuresults}
\end{center}
\end{table}

% \paragraph{Efficiency Analysis} 
\paragraph{Computational Efficiency}
% In addition to examining forecasting precision, we also evaluate computational performance. 
Fig. \ref{fig_sup:efficiency_all} (a-b) show the validation MAE curves of several competing STGNNs. Fig. \ref{fig_sup:efficiency_all} (c-d) detail the training speed, model size, and memory usage. And the numerical results of computational performance on \texttt{METR-LA} and \texttt{PEMS-BAY} data are shown in Tab. \ref{tab_sup:gpuresults}. 
It is worth noting that \gls{model} exhibits \textit{significantly faster training speed, with convergence achieved in just 10 minutes of GPU time}. Moreover, it has \textit{a smoother convergence curve, a lower error bound, and requires less memory consumption and smaller model size}, indicating high efficiency. 
The superiority is achieved by the combination of a technically simple structure, linear complexity, and parameter-efficient designs. Advanced STGNNs like \texttt{DGCRN} and \texttt{D2STGNN} apply alternating GNNs and temporal models, resulting in high computational burdens and unstable optimization. 
% Additional operations like self-attention or graph diffusion further complicate the process.

\subsection{Model Analysis}
This section provides deeper analysis on the model designs by conducting ablation studies and case studies.
\begin{table}[!htbp]
\caption{Ablation studies on \texttt{METR-LA} and \texttt{PEMS03} data.}\label{tab_ablation}
  \centering
  \setlength{\tabcolsep}{1pt}
  \begin{small}
  \renewcommand{\multirowsetup}{\centering}
  \resizebox{1\columnwidth}{!}{
  \begin{tabular}{c|c|c c c c|c c c c}
    % \cmidrule[1pt]{3-8}
    \toprule
     \multicolumn{2}{c}{\textsc{Datasets}} & \multicolumn{4}{c|}{\texttt{METR-LA}} & \multicolumn{4}{c}{\texttt{PEMS03}}  \\
     \cmidrule[0.7pt]{1-10}
    \multicolumn{2}{c|}{\textsc{Variations}}& \scalebox{0.92}{Full}& 
    \multicolumn{1}{c}{\scalebox{0.92}{w/o $\mathbf{E}_{T}$}} &
    \multicolumn{1}{c}{\scalebox{0.92}{w/ $\mathbf{A}_{\text{pre}}$}}  & \multicolumn{1}{c|}{\scalebox{0.92}{\makecell{w/o\\ \scriptsize \textsc{SpaceMixer}}}} & \scalebox{0.92}{Full} & \multicolumn{1}{c}{\scalebox{0.92}{w/o $\mathbf{E}_{T}$}} &
    \multicolumn{1}{c}{\scalebox{0.92}{w/ $\mathbf{A}_{\text{pre}}$}}  & \multicolumn{1}{c}{\scalebox{0.92}{\makecell{w/o \\\scriptsize \textsc{SpaceMixer}}}}\\
    \cmidrule[0.7pt]{1-10}
    \multicolumn{2}{c|}{\textsc{Metric}}  & \scalebox{0.92}{MAE}  & \scalebox{0.92}{MAE}  & \scalebox{0.92}{MAE}  & \scalebox{0.92}{MAE}  & \scalebox{0.92}{MAE}  & \scalebox{0.92}{MAE} & \scalebox{0.92}{MAE}  & \scalebox{0.92}{MAE}\\
    \midrule
     \multicolumn{2}{c|}{\scalebox{0.92}{15 min}} & \scalebox{0.92}{\textbf{2.66}}& \scalebox{0.92}{{2.88}} & \scalebox{0.92}{2.64} & \scalebox{0.92}{2.84} & \scalebox{0.92}{\textbf{13.15}}& \scalebox{0.92}{{14.85}} & \scalebox{0.92}{13.48} & \scalebox{0.92}{14.64}\\
     \multicolumn{2}{c|}{\scalebox{0.92}{30 min}} & \scalebox{0.92}{\textbf{2.99}}& \scalebox{0.92}{{3.32}}  & \scalebox{0.92}{2.98} & \scalebox{0.92}{3.23} &\scalebox{0.92}{\textbf{14.20}} & \scalebox{0.92}{{16.62}}  & \scalebox{0.92}{14.54} & \scalebox{0.92}{15.91}\\
     \multicolumn{2}{c|}{\scalebox{0.92}{60 min}}& \scalebox{0.92}{\textbf{3.36}}& \scalebox{0.92}{{3.76}} & \scalebox{0.92}{3.38}  & \scalebox{0.92}{3.65} &\scalebox{0.92}{\textbf{15.79}} & \scalebox{0.92}{{18.68}} & \scalebox{0.92}{15.95} & \scalebox{0.92}{17.45}\\
    \midrule
     \multicolumn{1}{c|}{\multirow{2}{*}{\scalebox{0.92}{\rotatebox{90}{Ave.}}}} 
     &\scalebox{0.92}{MAE} & \scalebox{0.92}{\textbf{2.95}}& \scalebox{0.92}{{3.25}} & \scalebox{0.92}{2.95}  & \scalebox{0.92}{3.18} & \scalebox{0.92}{\textbf{14.18}}& \scalebox{0.92}{{16.46}} & \scalebox{0.92}{14.45}  & \scalebox{0.92}{15.76}\\
     & \scalebox{0.92}{MSE} & \scalebox{0.92}{\textbf{35.35}}& \scalebox{0.92}{{39.97}} & \scalebox{0.92}{35.89} & \scalebox{0.92}{39.40} &\scalebox{0.92}{\textbf{603.29}} & \scalebox{0.92}{{766.64}} & \scalebox{0.92}{610.47} & \scalebox{0.92}{676.63}\\
    \bottomrule
  \end{tabular}}
  \end{small}
\end{table}
\paragraph{Ablation Study} 
Ablation studies are conducted to test the effectiveness of the model designs. 
We consider the following three variations:
\begin{itemize}
    \item w/ $\mathbf{A}_{\text{pre}}$: We add additional diffusion graph convolutions \cite{li2018diffusion} with three forward and backward diffusion steps using distance-based graphs in \texttt{SpaceMixer}.
    \item w/o \textsc{SpaceMixer}: We remove the ``\textsc{SpaceMixer}'' by keeping each channel independent \cite{nie2022time}. In this scenario, our model shares a structure similar to that of STID \cite{shao2022spatial}.
    \item w/o $\mathbf{E}_T$: We remove the spatiotemporal node embedding, and our model degrades into a spatiotemporal \textsc{Mlp-Mixer} model \cite{chen2023tsmixer}.
\end{itemize}

For each of them, we keep the same experimental settings in Section \ref{sec:exp} and report the forecasting results.
Table \ref{tab_ablation} presents the forecasting performance of different model variations.
% ``w/o $\mathbf{E}_T$'' indicates that the ST-contextualization is removed and the model degrades into a MLP-Mixer. ``w/ $\mathbf{A}_{\text{pre}}$'' incorporates predefined graphs with additional diffusion graph convolutions \cite{li2018diffusion}. ``w/o \textsc{SpaceMixer}'' means that we remove the \textsc{SpaceMixer} layer.
The results clearly justify the effectiveness of the model designs of NexuSQN. %and support our hypothesis.

{\paragraph{Spatiotemporal Embedding} The proposed STNE plays a critical role in improving forecasting performance. To understand the effects of temporal information in this embedding, we perform two ablation studies: (1) removing $\mathbf{U}$ from $\mathbb{E}_T$, and (2) incorporating $\mathbf{U}$ to the input of other baseline models. DCRNN and GWNET are selected for comparison (denoted with a superscript $^\dagger$.) Results in Tab. \ref{tab_ablation_sup} show that the temporal information is indispensable for improving the performance of NexuSQN. However, other baselines such as DCRNN and GWNET benefit little from the temporal encoding. This may be caused by the inefficacy in contextualizing spatiotemporal information in complicated architectures.}

\begin{table}[!htbp]
\caption{{Ablation studies on the spatiotemporal node embedding.}}\label{tab_ablation_sup}
  \centering
  \setlength{\tabcolsep}{1pt}
  \begin{small}
  \renewcommand{\multirowsetup}{\centering}
  \resizebox{1\columnwidth}{!}{
  \begin{tabular}{c|c|c c c c|c c c c}
    % \cmidrule[1pt]{3-8}
    \toprule
     \multicolumn{2}{c}{\textsc{Datasets}} & \multicolumn{4}{c|}{\texttt{METR-LA}} & \multicolumn{4}{c}{\texttt{PEMS03}}  \\
     \cmidrule[0.7pt]{1-10}
    \multicolumn{2}{c|}{\textsc{Variations}}& \scalebox{0.92}{Full}& 
    \multicolumn{1}{c}{\scalebox{0.92}{w/o $\mathbf{U}$}} &
    \multicolumn{1}{c}{\scalebox{0.92}{DCRNN$^\dagger$}}  & \multicolumn{1}{c|}{\scalebox{0.92}{GWNET$^\dagger$}} & \scalebox{0.92}{Full} & \multicolumn{1}{c}{\scalebox{0.92}{w/o $\mathbf{U}$}} &
    \multicolumn{1}{c}{\scalebox{0.92}{DCRNN$^\dagger$}}  & \multicolumn{1}{c}{\scalebox{0.92}{GWNET$^\dagger$}}\\
    \cmidrule[0.7pt]{1-10}
    \multicolumn{2}{c|}{\textsc{Metric}}  & \scalebox{0.92}{MAE}  & \scalebox{0.92}{MAE}  & \scalebox{0.92}{MAE}  & \scalebox{0.92}{MAE}  & \scalebox{0.92}{MAE}  & \scalebox{0.92}{MAE} & \scalebox{0.92}{MAE}  & \scalebox{0.92}{MAE}\\
    \midrule
     \multicolumn{2}{c|}{\scalebox{0.92}{15 min}} & \scalebox{0.92}{\textbf{2.66}}& \scalebox{0.92}{{2.78}} & \scalebox{0.92}{2.81} & \scalebox{0.92}{2.74} & \scalebox{0.92}{\textbf{13.15}}& \scalebox{0.92}{{13.42}} & \scalebox{0.92}{14.53} & \scalebox{0.92}{13.66}\\
     \multicolumn{2}{c|}{\scalebox{0.92}{30 min}} & \scalebox{0.92}{\textbf{2.99}}& \scalebox{0.92}{{3.12}}  & \scalebox{0.92}{3.25} & \scalebox{0.92}{3.13} &\scalebox{0.92}{\textbf{14.20}} & \scalebox{0.92}{{14.56}}  & \scalebox{0.92}{15.68} & \scalebox{0.92}{14.58}\\
     \multicolumn{2}{c|}{\scalebox{0.92}{60 min}}& \scalebox{0.92}{\textbf{3.36}}& \scalebox{0.92}{{3.45}} & \scalebox{0.92}{3.73}  & \scalebox{0.92}{3.60} &\scalebox{0.92}{\textbf{15.79}} & \scalebox{0.92}{{16.17}} & \scalebox{0.92}{18.20} & \scalebox{0.92}{16.23}\\
    \midrule
     \multicolumn{1}{c|}{\multirow{2}{*}{\scalebox{0.92}{\rotatebox{90}{Ave.}}}} 
     &\scalebox{0.92}{MAE} & \scalebox{0.92}{\textbf{2.95}}& \scalebox{0.92}{{3.07}} & \scalebox{0.92}{3.20}  & \scalebox{0.92}{ 3.09 } & \scalebox{0.92}{\textbf{14.18}}& \scalebox{0.92}{{14.51}} & \scalebox{0.92}{15.88}  & \scalebox{0.92}{14.65}\\
     & \scalebox{0.92}{MAPE} & \scalebox{0.92}{\textbf{8.04\%}}& \scalebox{0.92}{{8.47\%}} & \scalebox{0.92}{8.89\%} & \scalebox{0.92}{8.49\%} &\scalebox{0.92}{\textbf{13.77\%}} & \scalebox{0.92}{{14.43\%}} & \scalebox{0.92}{15.70\%} & \scalebox{0.92}{14.90\%}\\
    \bottomrule
  \end{tabular}}
  \end{small}
  \begin{tablenotes} 
% \tiny
\scriptsize
{
\item $^\dagger$ We equip baselines with time-related features, including the time of day and day of the week.
}
\end{tablenotes} 
\end{table}

\paragraph{Selection of Kernels} We evaluate the impacts of different kernel methods in Tab. \ref{tab_kernel}. As can be seen, the \texttt{softmax} kernel achieves the best performance, suggesting that the normalization and nonlinearity property of kernels can help to maintain the competitiveness of shallow \gls{model} with deep multi-layer STGNNs or Transformers.

\begin{table}[!htbp]
\caption{Performance of different kernels.}\label{tab_kernel}
  \centering
  \setlength{\tabcolsep}{2pt}
  \begin{small}
  \renewcommand{\multirowsetup}{\centering}
  \scalebox{0.9}{
  \begin{tabular}{c c| c c c c c}
    \toprule
     \multicolumn{2}{c}{\textsc{Datasets}} & \multicolumn{5}{c}{\texttt{PEMS08}}   \\
   \midrule
    \multicolumn{2}{c|}{\makecell{\textsc{Kernels}\\$\phi(x)=$}} & 
    \multicolumn{1}{c|}{\scalebox{0.92}{$\texttt{softmax}(x)$}} &
    \multicolumn{1}{c|}{\scalebox{0.92}{$\texttt{elu}(x)+1$}}  & \multicolumn{1}{c|}{\scalebox{0.92}{$\texttt{relu}(x/\vert x\vert)$}}& \multicolumn{1}{c|}{\scalebox{0.92}{$\texttt{exp}(x)$}}& \multicolumn{1}{c}{\scalebox{0.92}{$x/\vert x\vert$}}\\
    \midrule
    \multicolumn{2}{c|}{\textsc{Metric}}  & \scalebox{0.92}{MAE}  & \scalebox{0.92}{MAE}  & \scalebox{0.92}{MAE} & \scalebox{0.92}{MAE}& \scalebox{0.92}{MAE}\\
    \midrule
     \multicolumn{2}{c|}{\scalebox{0.92}{15 min}}  & \scalebox{0.92}{{13.17}} & \scalebox{0.92}{13.73} & \scalebox{0.92}{13.28} & \scalebox{0.92}{13.79} & \scalebox{0.92}{13.30}\\
     \multicolumn{2}{c|}{\scalebox{0.92}{30 min}} & \scalebox{0.92}{{13.96}}  & \scalebox{0.92}{14.75} & \scalebox{0.92}{14.05} & \scalebox{0.92}{14.81}& \scalebox{0.92}{14.12}\\
     \multicolumn{2}{c|}{\scalebox{0.92}{60 min}} & \scalebox{0.92}{{15.28}} & \scalebox{0.92}{16.56}  & \scalebox{0.92}{15.38} & \scalebox{0.92}{16.68}& \scalebox{0.92}{15.53}\\
    \midrule
     % \multicolumn{1}{c|}{\multirow{2}{*}{\scalebox{0.92}{\rotatebox{90}{Ave.}}}} 
     % &\scalebox{0.92}{MAE} & \scalebox{0.92}{{13.98}} & \scalebox{0.92}{14.79}  & \scalebox{0.92}{14.08} & \scalebox{0.92}{14.87}& \scalebox{0.92}{14.15}\\
     % & \scalebox{0.92}{MAPE}  & \scalebox{0.92}{{9.02}} & \scalebox{0.92}{9.55} & \scalebox{0.92}{9.14} & \scalebox{0.92}{9.68}& \scalebox{0.92}{9.23}\\
     &\scalebox{0.92}{Ave.} & \scalebox{0.92}{{13.98}} & \scalebox{0.92}{14.79}  & \scalebox{0.92}{14.08} & \scalebox{0.92}{14.87}& \scalebox{0.92}{14.15}\\
    \bottomrule
  \end{tabular}}
  \end{small}
\end{table}

\begin{figure}[!htbp]
    \centering
    \includegraphics[width=0.95\columnwidth]{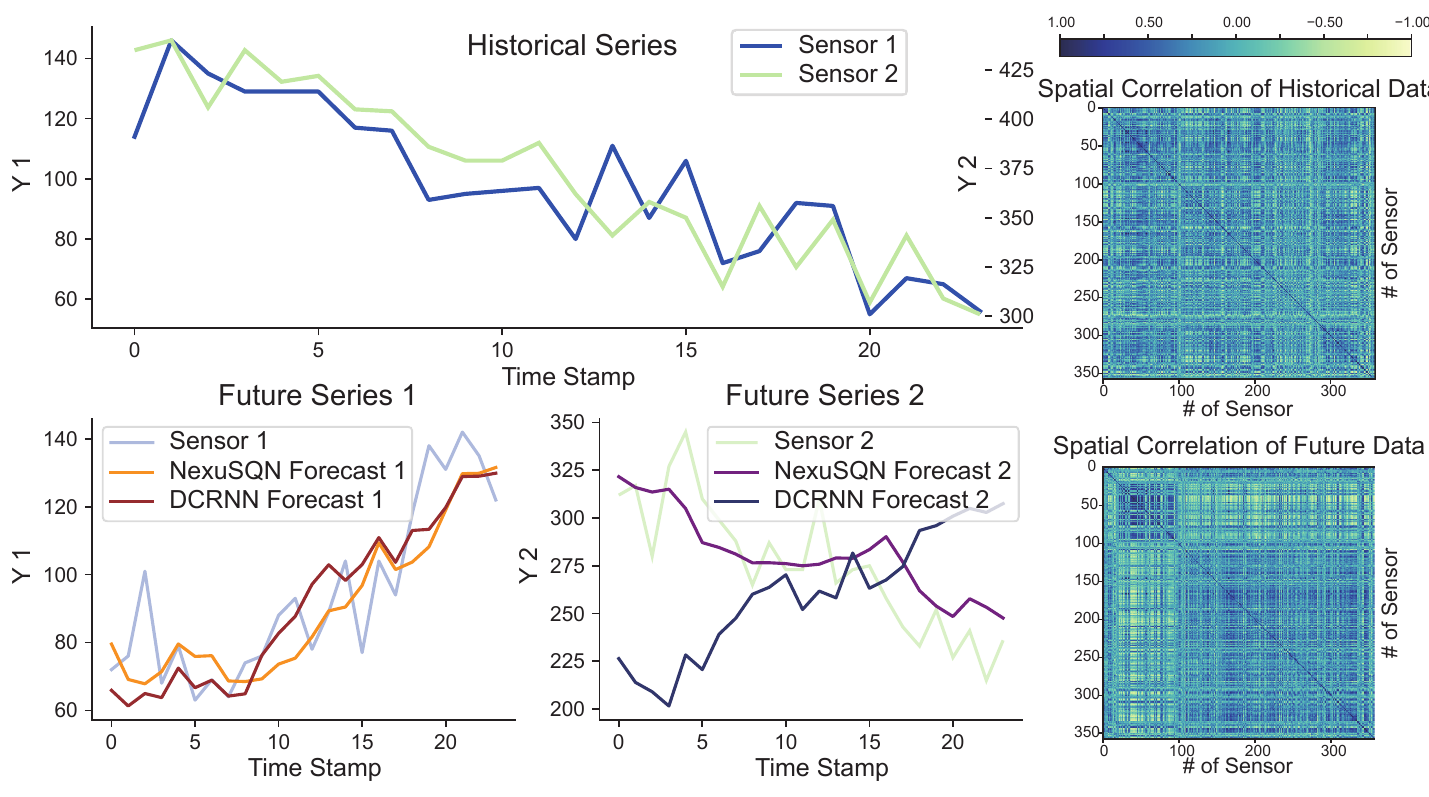}
    \caption{Illustration of spatial contextualization issue.}
    \label{case_spatial}
\end{figure}

\begin{figure}[!htbp]
    \centering
    \includegraphics[width=0.95\columnwidth]{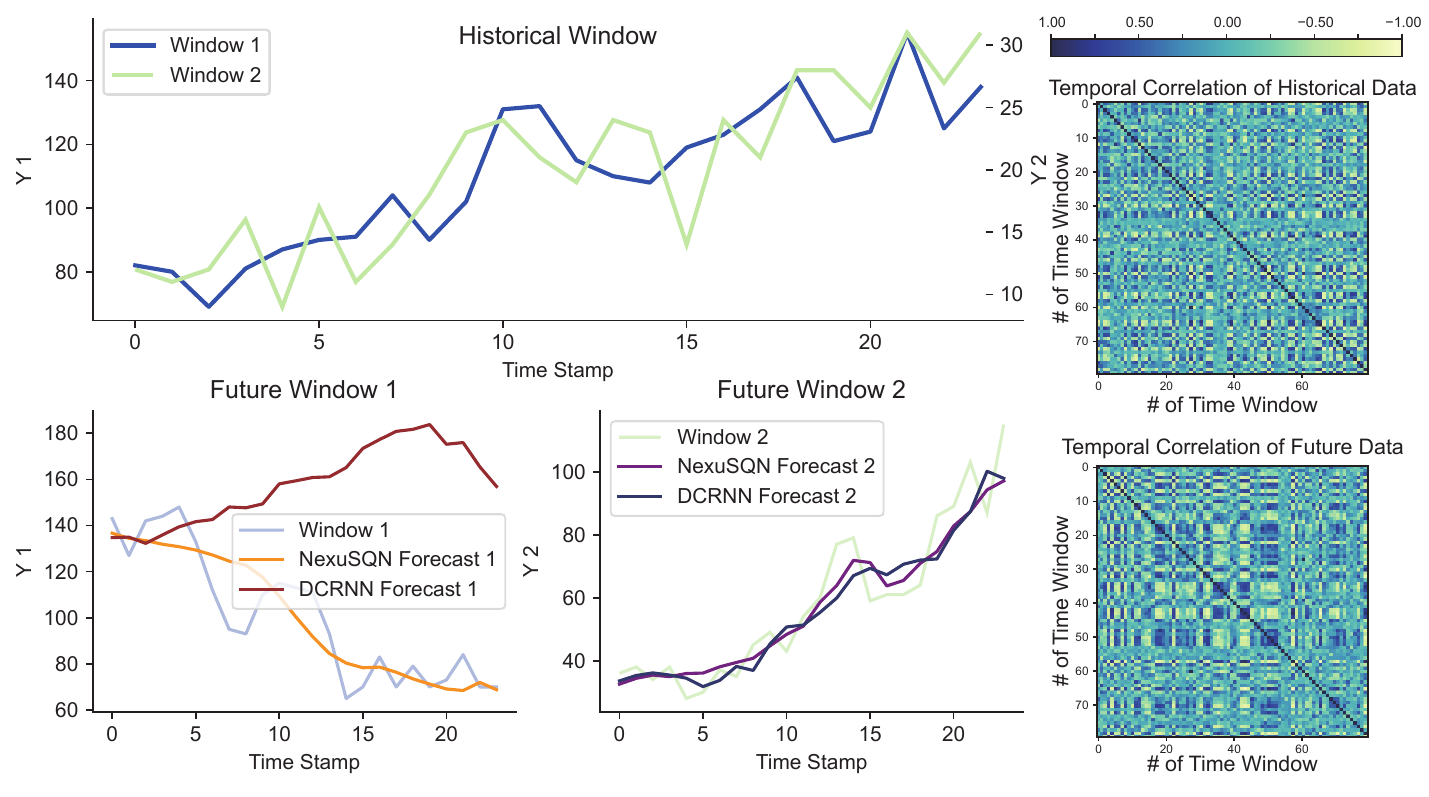}
    \caption{Temporal contextualization issue.}
    \label{fig:case_temporal}
\end{figure}

\paragraph{Spatiotemporal Contextualization}
We provide an example of the spatial contextualization effect in Fig. \ref{case_spatial}. An illustration of the temporal contextualization issue and the behavior of different models are also provided in Fig. \ref{fig:case_temporal}. As can be seen, the correlation of multivariate series can show distinct patterns between historical and future windows. With the design of spatiotemporally contextualized mixers, our model can distinguish different short-term patterns in both space and time and forecast the true trend.

\subsection{{Additional Benefits of Parsimonious Models}}
% Real-world applications favor simple models over complex ones. 
Based on the minimalist MLP-Mixer structure, \acrshort{model} features both accuracy, efficiency and flexibility.
The additional benefits of a parsimonious model are evident in the following paragraphs.

\begin{figure}[!htbp]
    \centering
    \includegraphics[width=\columnwidth]{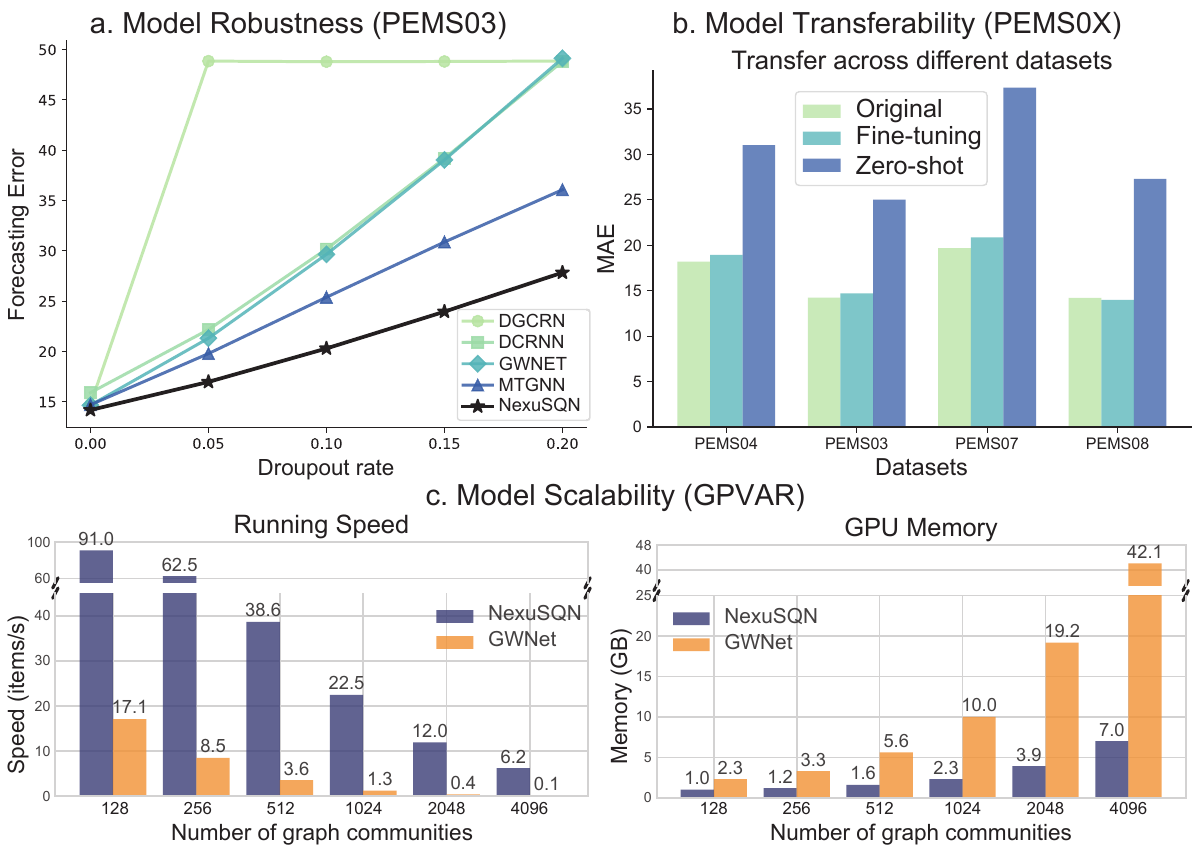}
    \caption{Robustness, transferability, and scalability of NexuSQN.}
    \label{fig:benefits}
\end{figure}

\paragraph{Robustness Under Attack} 
To evaluate the robustness under model attack, we select several representative STGNNs and randomly corrupt a proportion of linear weights in the input embedding or the final readout layer. These corrupted values are filled with zeros.
Fig. \ref{fig:benefits} (a) displays the forecasting errors after randomly corrupting some proportions of weight parameters in the input or readout layer.
As can be seen, \gls{model} is more robust than complicated ones when faced with external attack.

\paragraph{Flexibility to Transfer} 
As suggested in \cite{cini2023taming}, node embedding can facilitate transfer between different datasets. We first pretrain our model on small \texttt{PEMS08} data, then freeze all parameters except node embedding. During the transfer stage, we randomly initialize the node embedding according to the spatial dimension of the target domains (i.e., \texttt{PEMS03}, \texttt{PEMS04}, and \texttt{PEMS07}) and fine tune the embedding parameter.
\gls{model} can easily be transferred by resorting to the learnable node embedding. In Fig. \ref{fig:benefits} (b), we train \gls{model} on \texttt{PEMS08} and only fine-tune the node embedding on other data. We find that the fine-tuning of embedding performs comparably with the models trained from scratch.

\paragraph{Scalability}
Fig \ref{fig:benefits} (c) shows the scalability of models with different numbers of graph communities using the synthetic GPVAR data \cite{zambon2022az}. A larger number of graph communities have more numbers of nodes.
Our model approximates to scale linearly to the data size in space and time load, showing greater scalability than STGNN such as GWNet.

\section{{Deployed Applications in Urban Networks}}\label{sec:deployed}
Existing models for \acrshort{task} have been extensively validated on public benchmarks, but their effectiveness in large-scale urban road networks and real-world production environments has been less studied.
Therefore, we deploy our model in real-world challenges and showcase its practical applications in two testing stages. 
% \textbf{Online demonstrations are provided on the platform}: \underline{\url{http://110.42.248.92/index.php/traffic-forecasting/}}.

% \subsection{Deployment on Large-Scale Urban Networks}
\subsection{Stage 1: Scalability in Offline Pre-training}
The scalability of the proposed \gls{model} was tested in an Urban Congestion Project (UCP). 
% The UCP aims to model the spatiotemporal evolution of traffic congestion in urban road networks using data-driven methods. 
In UCP, a vast database containing 27,070 taxis with more than 900 million vehicle trajectories was collected and desensitized. This database was collected from various taxi companies, covering the primary road networks of Beijing, Shanghai, and Shenzhen cities in China. The raw trajectory database was processed using the map-matching algorithm \cite{brakatsoulas2005map} and the average speeds were calculated. 
We then aggregated the speed values into three grid-based datasets with a 300-m resolution in Tab. \ref{tab_large_datasets}.
% The speed values were further aggregated by a 300-m resolution to construct three grid-based datasets in Tab. \ref{tab_large_datasets}. 

\begin{table}[!htbp]
\centering
\caption{Statistics of private urban traffic datasets.}
\label{tab_large_datasets}
\small
\centering
\begin{small}
\setlength{\tabcolsep}{1pt}
\resizebox{1\columnwidth}{!}{
\begin{tabular}{c| c| c| c| c}
\toprule
 \multicolumn{1}{c|}{\textsc{Datasets}} & \textsc{Range} & \textsc{Data points} & \textsc{Grids}  & \textsc{Vehicles}\\
\midrule
\texttt{Beijing} &115.7°-117.4° E, 39.4°-41.6° N  & 270 million &  8,267& 12,060 \\
\texttt{Shenzhen} &113.5°-114.4° E, 22.3°-22.5° N  & 210 million & 7,680 & 6,330  \\
\texttt{Shanghai} & 120.5°-122.1° E, 30.4°-31.5° N  & 420 million & 8,854 & 8,680  \\
\bottomrule
\end{tabular}}
\end{small}
\end{table}

% TODO: how many cells? how large? [done]

% UCP data contains more than $60\%$ of missing data (see Fig. \ref{fig:traffic_network_true}), 

% making it inconvenient to investigate the temporal evolution and spatial propagation of traffic congestion at the regional level. 

\begin{figure}[!htbp]
\centering
\subfigure[Traffic forecast at network level (Beijing and Shanghai)]{
\centering
\includegraphics[scale=0.15]{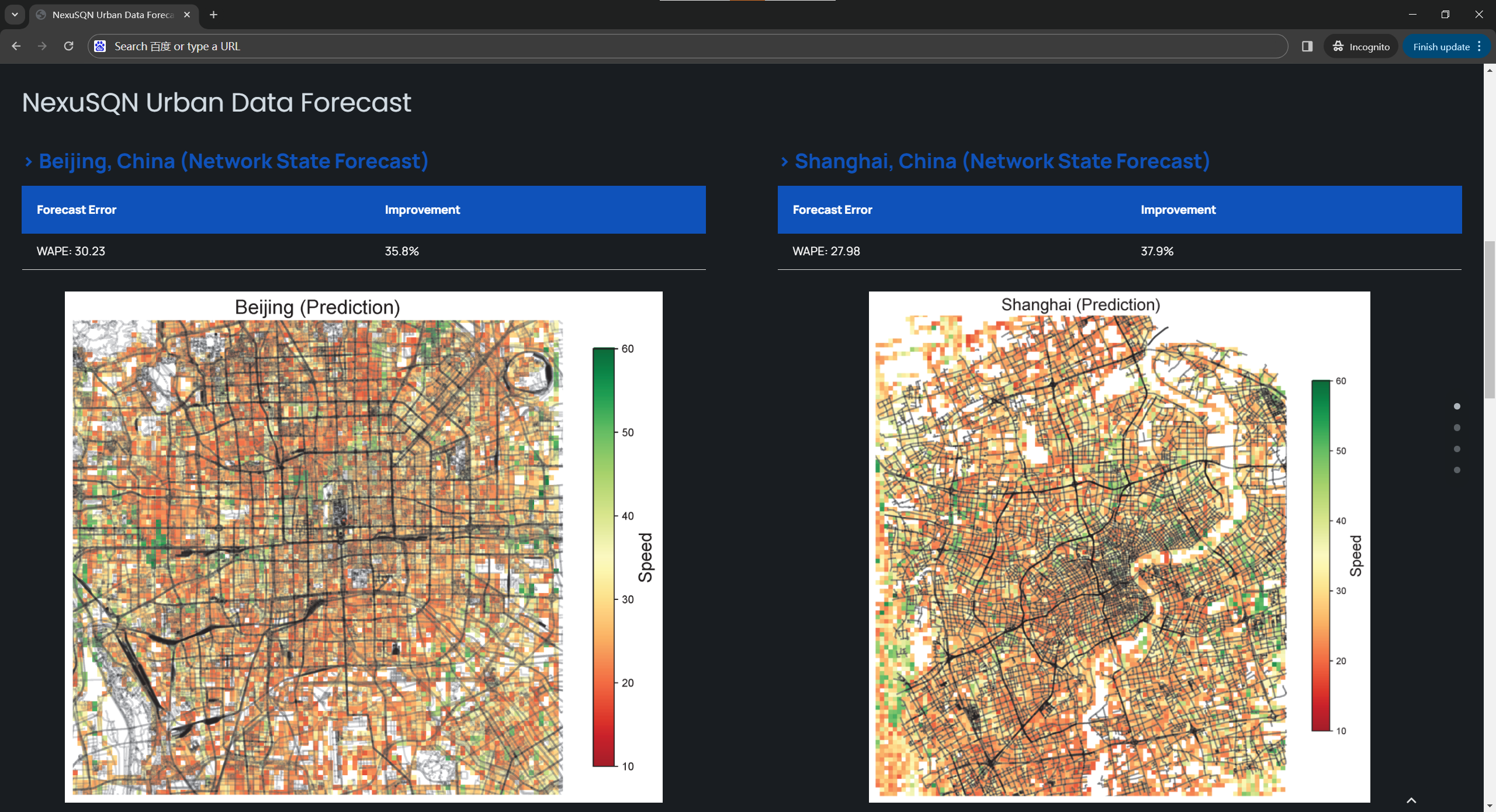}}
\subfigure[Traffic forecast at regional level (Shanghai)]{
\centering
\includegraphics[scale=0.15]{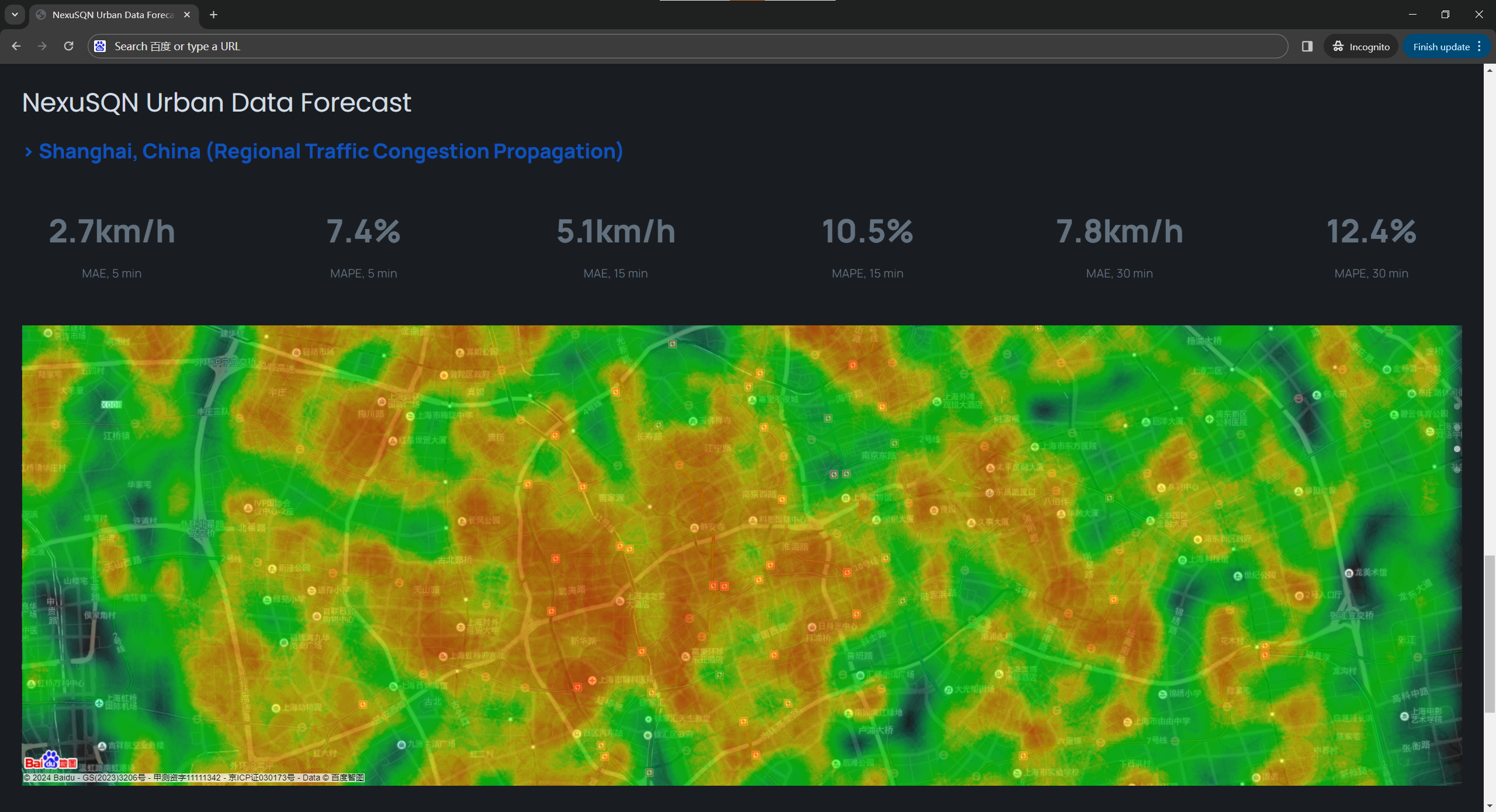}}
\caption{Scalability demonstrations.}
\label{fig:dashboard_1}
\end{figure}

To investigate the temporal evolution and spatial propagation of traffic congestion at the regional level, deep learning models were trained on an offline server.
These models used the three large-scale urban datasets to forecast the traffic speed across the entire network.
% to forecast the network-wide traffic speed in the three large-scale urban datasets. 
% Speed forecasts are shown in Fig. \ref{fig:traffic_network}, and the full evaluation results are given in Tab. \ref{tab:baidu_results}.
The regional and network-level forecasts of \gls{model} deployed on our internal server are shown in Fig. \ref{fig:dashboard_1}. Full quantitive results of the baselines are given in Tab. \ref{tab:baidu_results}.
% Tab. \ref{tab:baidu_results}.
% Forecasting long-term and large-scale urban traffic with a large proportion of missing data is a challenge for 
Many advanced STGNNs and Transformers adopt computationally intensive techniques. 
% that are developed primarily in highway traffic systems. 
% The behaviors of traffic flows in urban networks are quite different and more complex than those of highway systems. Structural priors in STGNNs, such as proximity-based adjacency graphs, may be ineffective.
As a consequence, the high dimensionality of urban networks made many of them fail to respond promptly on a single computing server. In contrast, \gls{model} completed the pre-training task with high efficiency and achieved greater accuracy ($\geq 35\%$ improvement in WAPE) than other baselines.
% Importantly, forecast results can also be adopted to reveal the large-scale spatiotemporal evolution process of traffic congestion.
% , as shown in the online platform in Fig. \ref{fig:dashboard_1}. 

\begin{table}[!htbp]
\caption{Results on \texttt{Baoding}, \texttt{Yizhuang}, \texttt{Beijing}, \texttt{Shenzhen} and \texttt{Shanghai}  urban traffic data. (192 to 192)}
\small
\setlength{\tabcolsep}{2pt}
\begin{center}
\resizebox{1\columnwidth}{!}{
\begin{tabular}{l | cc| cc| cc| cc| cc }
\cmidrule[1pt]{2-11}
 \multicolumn{1}{c}{} & \multicolumn{2}{c|}{\texttt{Baoding}} & \multicolumn{2}{c|}{\texttt{Yizhuang}}  & \multicolumn{2}{c|}{\texttt{Beijing}} & \multicolumn{2}{c|}{\texttt{Shenzhen}} & \multicolumn{2}{c}{\texttt{Shanghai}}\\
\cmidrule[0.5pt]{2-11}
   \multicolumn{1}{c}{} & \scriptsize MAE & \scriptsize WAPE & \scriptsize MAE & \scriptsize WAPE   & \scriptsize MAE & \scriptsize WAPE  & \scriptsize MAE & \scriptsize WAPE  & \scriptsize MAE & \scriptsize WAPE  \\
 \midrule
{MTGNN} &  13.66 & 59.55 & 16.41 & 77.20 & OOM& OOM& 22.38&55.91 & OOM&OOM \\
{AGCRN} & OOM  & OOM & 12.24 & 59.61 & OOM& OOM&OOM &OOM & OOM&OOM \\
{StemGNN} &  12.34 & 55.59 & 12.97 & 61.16 &OOM & OOM&OOM & OOM& OOM&OOM \\
{TimesNet} & 10.12  & 48.77 &11.78 & 56.60  &15.52 & 45.69 & 18.22&45.37 & 16.39 & 40.88\\
{Pyraformer} &  8.42 & 37.92 &10.86  & 51.05  &OOM &OOM &OOM &OOM &OOM &OOM\\
{Autoformer} & 9.09  & 40.94 & 11.55 &  54.31 & 17.47 & 47.11& 21.22& 52.67& 17.75&45.07\\
{FEDformer} & 8.44  & 37.87 &10.06 & 47.33 & 17.69 & 47.72 & 22.25 & 55.21 & 18.10 &45.96 \\
{Informer} & 8.36  & 37.68 & 10.91 &  49.59 & 17.37 & 46.83 &21.23 &52.71 & 17.09&43.39\\
{DLinear} & 8.65  & 38.76 & 10.71 & 50.36 & 13.69 & 36.92 & 13.61& 37.79 &13.48 &34.24\\
{PatchTST} & 9.99  & 44.98 &10.53 &49.52  &13.62 &36.74 & 14.55 &36.11 & 12.95&32.89 \\
\midrule
\textbf{\gls{model}} & \textbf{7.89}  & \textbf{35.54}  & \textbf{9.55}  &   \textbf{44.91}  & \textbf{11.21} & \textbf{30.23}& \textbf{10.89} & \textbf{32.01} & \textbf{11.03}& \textbf{27.98}\\
\bottomrule
\end{tabular}}
\label{tab:baidu_results}
\end{center}
\end{table}

\subsection{Stage 2: Feasibility in Online Parallel Testing}
\begin{figure}[!htbp]
\centering
\subfigure[Online testing system architecture and results.]{
\includegraphics[width=0.95\linewidth]{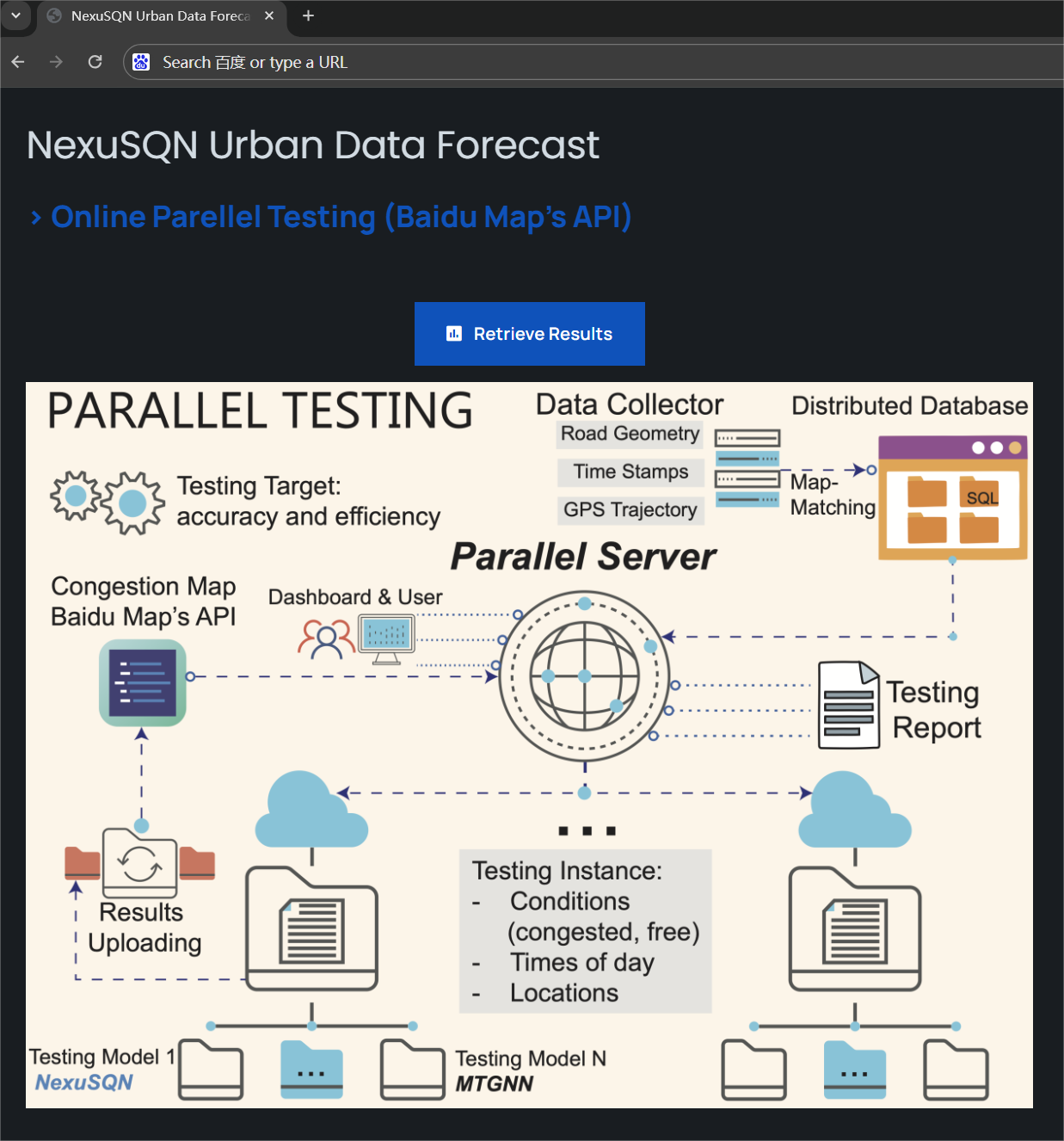}}
\subfigure[Online testing results of NexuSQN.]{
\includegraphics[width=0.95\linewidth]{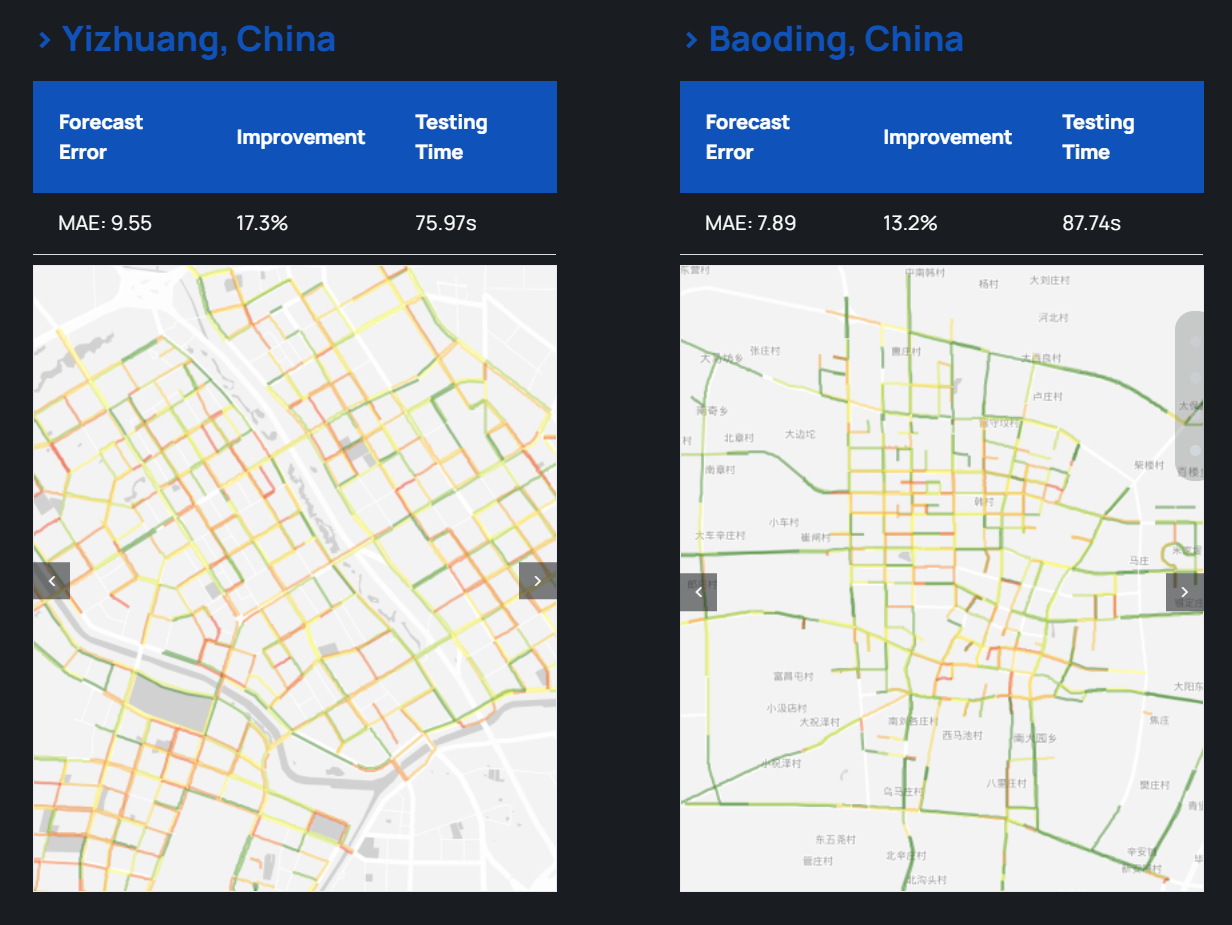}}
\caption{Online parallel testing.}
\label{fig:parallel_test}
\end{figure}

In this testing phase, we collaborated with Baidu to conduct an online parallel test in Baoding and Yizhuang districts. 
The aim of the test was to examine the feasibility of the models in generating fine-grained real-time congestion maps in real-world production environments.
% We perform a parallel test using a central server and Baidu Map's API. 
The architecture of the system is shown in Fig. \ref{fig:parallel_test} (a).

For each test instance, we fine-tuned pretrained models from stage 1 using data from the first month of 2023 in Baoding and Yizhuang datasets. Then the online data streaming was synchronous for all instances and model inference was carried out in each parallel computing server. 
To meet the requirement of real-time query using Baidu Map's API, traffic data flow was updated every 3 minutes and models need to forecast the online stream.
The test results were displayed on the online dashboard, as shown in Fig. \ref{fig:parallel_test} (b). Quantitative results are also given in Tab. \ref{tab:baidu_results}.
With high data throughput ($> 25$ batch/s), \gls{model} generated results in real time, e.g., 87.7 s in Baoding data. 
Due to the delay in data transmission, many STGNNs with high complexity could not complete the real-time testing task in limited execution times.
Compared to other testing models like Autoformer, \acrshort{model} showed significant improvements. It achieved over 15\% higher accuracy in generating traffic state maps and over 20 times faster online inference speed.
% can generate the traffic state map with over $15\%$ improvement in accuracy and over 20x acceleration in online inference speed.
Overall, the testing phase demonstrated the efficiency and effectiveness of our model in production environments.

% \subsection{Additional Results on Private data} 
% The models are trained or fine-tuned to forecast the future 192 steps using the historical 192 steps. 
% Online parallel testing results are given in Fig. \ref{fig:parallel_test_2}.
% \begin{figure}[!htbp]
% \centering
% \includegraphics[width=1\linewidth]{figs/dashboard_baidu2.png}
% \caption{Online testing results of NexuSQN.}
% \label{fig:parallel_test_2}
% \end{figure}
% As a reference, the ground-truth speed maps of the Beijing and Shanghai districts are also displayed in Fig. \ref{fig:traffic_network_true}.

{\section{Conclusion and Outlook}\label{sec:conclusion}
This paper extends MLP-Mixers to large-scale urban traffic data forecasting problems. By addressing the spatiotemporal contextualization issue, we demonstrate a scalable spatiotemporal MLP-Mixer called NexuSQN. NexuSQN customizes structured time and space mixing operations with learnable embedding. Extensive evaluations in traffic datasets show that NexuSQN performs comparably or even better than existing advanced STGNNs and Transformers in accuracy, efficiency, scalability, robustness, and transferability. 
It also has the potential to predict general spatiotemporal data in urban systems. 
We also deploy it in a real-world urban congestion project and present its practical applications.
Future work could deploy it on edge computing devices and achieve online model training. It could also be combined with distributed techniques to achieve privacy-preserving purposes.
}

\appendix
In this technical appendix, we provide additional details on experimental setups and model implementations as a supplement to the main paper. In addition, supplementary evaluation results and case studies are provided for additional discussion.

\begin{table}[!htbp]
\centering
% \begin{minipage}[c]{0.49\textwidth}
\centering
\caption{Statistics of public urban computing benchmarks.}
\label{tab_datasets_sup}
\small
\centering
\begin{small}
\setlength{\tabcolsep}{1pt}
\resizebox{1\columnwidth}{!}{
\begin{tabular}{l|c| c| c| c| c| c}
\toprule
 \multicolumn{2}{c|}{\textsc{Datasets}} & \textsc{Type} & \textsc{Steps} & \textsc{Nodes} & \textsc{Edges} & \textsc{Interval}\\
% \midrule
% \multirow{8}{*}{\rotatebox{90}{Traffic}} &
% \texttt{METR-LA} &  speed & 34,272 & 207 & 1,515 & 5 min \\
% & \texttt{PEMS-BAY} &  speed & 52,128 & 325 & 2,369 & 5 min \\
% % \midrule
% & \texttt{PEMS03} &  volume & 26,208 & 358 & 546 & 5 min \\
% & \texttt{PEMS04} &  volume & 16,992 & 307 & 340 & 5 min \\
% & \texttt{PEMS07} &  volume & 28,224 & 883 & 866 & 5 min \\
% & \texttt{PEMS08} &  volume & 17,856 & 170 & 277 & 5 min \\
% % \midrule
% & \texttt{TrafficL} & occupancy & 17,544 & 862 & - & 60 min \\
% % \midrule
% & \texttt{LargeST-GLA} & volume & 525,888 & 3,834 & 98,703 & 15 min \\
\midrule
\multirow{3}{*}{\rotatebox{90}{Energy}} &
\texttt{Electricity} & electricity & 26,304 & 321 & - & 60 min \\
& \texttt{CER-EN} &  smart meters & 52,560 & 6,435 & 639,369 & 10 min \\
& \texttt{PV-US} &  solar power & 52,560 & 5,016 & 417,199 & 10 min \\
\midrule
\multirow{3}{*}{\rotatebox{90}{Environ.}} &
\texttt{AQI} &  pollutant & 8,760 & 437 & 2,699 & 60 min \\
& \texttt{Global Temp} & temperature & 17,544 & 3,850 & - & 60 min \\
& \texttt{Global Wind} &  wind speed & 17,544 & 3,850 & - & 60 min \\
\bottomrule
\end{tabular}}
\end{small}
% \end{minipage}
\end{table}

\subsection{Additional Dataset Descriptions}\label{appendix_data}
In addition to the traffic datasets discussed in the main paper, we also adopt several public urban computing datasets in other areas to evaluate our model, as shown in Tab. \ref{tab_datasets_sup}. This part provides detailed information about these datasets. Similarly, the input-output settings are given in Tab. \ref{tab_settings_appendix}.

\paragraph{Air Quality Benchmarks} \texttt{AQI} data from the Urban Air project \cite{zheng2015forecasting} record PM2.5 pollutant measurements collected by 437 air quality monitoring stations across 43 Chinese cities from May 2014 to April 2015 with an aggregation interval of 1 hour. Note that \texttt{AQI} data contains nearly $26\%$ missing data.

\paragraph{Energy Production and Consumption Benchmarks} Large-scale \texttt{PU-VS} production data \cite{hummon2012sub} consists of a simulated energy production by 5016 PV farms in the United States during 2006. The original observations are aggregated into a 30-minute window. \texttt{Electricity} benchmark is widely adopted to evaluate long-term forecast performance. It records load profiles (in kWh) measured hourly by 321 sensors from 2012 to 2014.
\texttt{CER-En}: smart meters measuring energy consumption from the Irish Commission for Energy Regulation Smart Metering Project \footnote{\url{https://www.ucd.ie/issda/data/commissionforenergyregulationcer}}. Following \cite{cini2022scalable}, we consider the full sensor network containing 6435 smart meters with a 30-minute aggregation interval.

\paragraph{Global Meteorological Benchmarks}
\texttt{Global Wind} and \texttt{Global Temp} \cite{wu2023interpretable} contain the hourly averaged wind speed and hourly temperature of meteorological 3850 stations around the world from 1 January 2019 to 31 December 2020 from the National Centers for Environmental Information (NCEI) system.

\begin{table}[!htbp]
\centering
\caption{Input and output settings.}
\label{tab_settings_appendix}
\small
\centering
\begin{small}
\setlength{\tabcolsep}{5pt}
\resizebox{0.8\columnwidth}{!}{
\begin{tabular}{l|c| c| c| c}
\toprule
 \multicolumn{2}{c|}{\textsc{Datasets}} & \textsc{window} & \textsc{Horizon} & \textsc{Graphs} \\
\midrule
\multirow{3}{*}{\rotatebox{90}{Energy}} &
\texttt{Electricity} & 336 & 96 & False \\
& \texttt{CER-EN} &  36 & 22 & True  \\
& \texttt{PV-US} &  36 & 22 & True \\
\midrule
\multirow{3}{*}{\rotatebox{90}{Environ.}} &
\texttt{AQI} &  24 & 3 & True \\
& \texttt{Global Temp} & 48 & 24 & False  \\
& \texttt{Global Wind} &  48 & 24 & False  \\
\bottomrule
\end{tabular}}
\end{small}
\end{table}

\subsection{Results on Energy Benchmarks}

\begin{table}[!htbp]
\caption{Long-term results on \texttt{Electricity} (in: 336, out: 96).}\label{tab_elec}
  \centering
  \begin{small}
  \renewcommand{\multirowsetup}{\centering}
  \setlength{\tabcolsep}{5pt}
 \resizebox{0.8\columnwidth}{!}{
  \begin{tabular}{c| c c |c c }
    % \cmidrule[1pt]{2-7}
    \toprule
     \multicolumn{1}{c}{\textsc{Dataset}} & \multicolumn{4}{c}{\texttt{Electricity}}  \\
     \midrule
    \multicolumn{1}{c|}{\textsc{Models}} & \multicolumn{1}{c}{\scalebox{0.8}{\textbf{\gls{model}}}}  & \multicolumn{1}{c|}{\scalebox{0.8}{DLinear}} &
    \multicolumn{1}{c}{\scalebox{0.8}{TimesNet}}  & \multicolumn{1}{c}{\scalebox{0.8}{SCINet}}\\
    % \cmidrule[0.7pt]{2-7}
    \midrule
    \multicolumn{1}{c|}{\textsc{Metric}}  & \scalebox{0.92}{MAE}  & \scalebox{0.92}{MAE}  & \scalebox{0.92}{MAE}  & \scalebox{0.92}{MAE} \\
    \midrule
    \multicolumn{1}{c|}{\scalebox{0.92}{24}} & \scalebox{0.92}{\textbf{178.87}} & \scalebox{0.92}{{185.35}} & \scalebox{0.92}{269.67} & \scalebox{0.92}{197.77}\\
    \multicolumn{1}{c|}{\scalebox{0.92}{48}} &  \scalebox{0.92}{\textbf{209.41}} & \scalebox{0.92}{{215.58}}  & \scalebox{0.92}{276.89} & \scalebox{0.92}{241.24}\\
    \multicolumn{1}{c|}{\scalebox{0.92}{72}} &  \scalebox{0.92}{\textbf{225.33}} & \scalebox{0.92}{{228.43}} & \scalebox{0.92}{283.32} & \scalebox{0.92}{246.48}\\
    \multicolumn{1}{c|}{\scalebox{0.92}{96}} &  \scalebox{0.92}{\textbf{234.89}} & \scalebox{0.92}{{236.07}} & \scalebox{0.92}{293.13}  & \scalebox{0.92}{252.44}\\
    % \cmidrule(lr){2-7}
    \midrule
     \scalebox{0.92}{MAE} & \scalebox{0.92}{\textbf{206.34}} & \scalebox{0.92}{{212.25}} & \scalebox{0.92}{276.86}  & \scalebox{0.92}{225.40}\\
    \scalebox{0.92}{MRE} & \scalebox{0.92}{\textbf{7.66}} & \scalebox{0.92}{{7.89}} & \scalebox{0.92}{10.29} & \scalebox{0.92}{8.38}\\
     \scalebox{0.92}{MAPE} & \scalebox{0.92}{\textbf{12.97}} & \scalebox{0.92}{{15.05}} & \scalebox{0.92}{15.53}  & \scalebox{0.92}{18.67}\\
    \bottomrule
  \end{tabular}}
  \end{small}
\end{table}

\begin{table}[!htbp]
\caption{Large-scale results on \texttt{CER-En} (in: 36, out: 22).}\label{tab_cer}
  \centering
  \begin{small}
  \setlength{\tabcolsep}{8pt}
  \renewcommand{\multirowsetup}{\centering}
  \resizebox{0.8\columnwidth}{!}{
  \begin{tabular}{c| c c| c c }
    % \cmidrule[1pt]{2-7}
    \toprule
\multicolumn{1}{c}{\textsc{Dataset}} & \multicolumn{4}{c}{\texttt{CER-En}}  \\
    \midrule
    \multicolumn{1}{c|}{\textsc{Models}} & \multicolumn{1}{c}{\scalebox{0.8}{\textbf{\gls{model}}}} & \multicolumn{1}{c|}{\scalebox{0.8}{TSMixer}} & \multicolumn{1}{c}{\scalebox{0.8}{GWNet}} &
    \multicolumn{1}{c}{\scalebox{0.8}{DCRNN}}  \\
    % \cmidrule[0.7pt]{2-7}
    \midrule
    \multicolumn{1}{c|}{Metric}  & \scalebox{0.92}{MAE}  & \scalebox{0.92}{MAE}  & \scalebox{0.92}{MAE}  & \scalebox{0.92}{MAE} \\
    \midrule
      \multicolumn{1}{c|}{\scalebox{0.92}{0.5 h}} & \scalebox{0.92}{\textbf{0.21}} & \scalebox{0.92}{0.23} & \scalebox{0.92}{{0.23}} & \scalebox{0.92}{0.22} \\
    \multicolumn{1}{c|}{\scalebox{0.92}{7.5 h}} &  \scalebox{0.92}{\textbf{0.26}} & \scalebox{0.92}{0.30} & \scalebox{0.92}{{0.34}}  & \scalebox{0.92}{0.28} \\
   \multicolumn{1}{c|}{\scalebox{0.92}{11 h}} &  \scalebox{0.92}{\textbf{0.27}} & \scalebox{0.92}{0.31} & \scalebox{0.92}{{0.35}} & \scalebox{0.92}{0.28} \\
    % \cmidrule(lr){2-7}
    \midrule
    \scalebox{0.92}{MAE} & \scalebox{0.92}{\textbf{0.25}} & \scalebox{0.92}{0.29} & \scalebox{0.92}{{0.31}} & \scalebox{0.92}{0.27}  \\
    \scalebox{0.92}{MRE} & \scalebox{0.92}{\textbf{0.42}}& \scalebox{0.92}{0.49} & \scalebox{0.92}{{0.51}} & \scalebox{0.92}{0.45} \\
    \scalebox{0.92}{MSE} & \scalebox{0.92}{\textbf{0.37}}& \scalebox{0.92}{0.49} & \scalebox{0.92}{{0.59}} & \scalebox{0.92}{0.46}  \\
    \bottomrule
  \end{tabular}}
  \end{small}
\end{table}

\begin{table}[!htbp]
\caption{Large-scale results on \texttt{PU-VS} (in: 36, out: 22).}\label{tab:pv_results}
  \centering
  \begin{small}
  \renewcommand{\multirowsetup}{\centering}
  \setlength{\tabcolsep}{1pt}
  \scalebox{0.85}{
  \begin{tabular}{c| c c c| c c c c c c}
    \toprule
     \multicolumn{1}{c}{\textsc{Dataset}} & \multicolumn{8}{c}{\texttt{PV-US}}  \\
     \midrule
    \multicolumn{1}{c|}{\textsc{Models}} & \multicolumn{1}{c}{\scalebox{0.8}{\textbf{\gls{model}}}} & \multicolumn{1}{c}{\scalebox{0.8}{DLinear}} & \multicolumn{1}{c|}{\scalebox{0.8}{TSMixer}} & \multicolumn{1}{c}{\scalebox{0.8}{DCRNN}} &
    \multicolumn{1}{c}{\scalebox{0.8}{GWNet}}  & \multicolumn{1}{c}{\scalebox{0.8}{GatedGN}} & \multicolumn{1}{c}{\scalebox{0.8}{GRUGCN}}& \multicolumn{1}{c}{\scalebox{0.8}{STGCN}}& \multicolumn{1}{c}{\scalebox{0.8}{MTGNN}}\\
    \midrule
    \multicolumn{1}{c|}{\textsc{Metric}}  & \scalebox{0.92}{MAE}  & \scalebox{0.92}{MAE}  & \scalebox{0.92}{MAE}  & \scalebox{0.92}{MAE}& \scalebox{0.92}{MAE}& \scalebox{0.92}{MAE}& \scalebox{0.92}{MAE}& \scalebox{0.92}{MAE}& \scalebox{0.92}{MAE} \\
    \midrule
    \multicolumn{1}{c|}{\scalebox{0.92}{0.5 h}} & \scalebox{0.92}{\textbf{1.30}}&\scalebox{0.92}{1.59} &\scalebox{0.92}{1.80}&\scalebox{0.92}{1.47} & \scalebox{0.92}{{1.65}} & \scalebox{0.92}{1.48} & \scalebox{0.92}{1.61} &\scalebox{0.92}{1.44} &\scalebox{0.92}{1.88} \\
    \multicolumn{1}{c|}{\scalebox{0.92}{7.5 h}} &  \scalebox{0.92}{\textbf{2.64}} &\scalebox{0.92}{3.11} &\scalebox{0.92}{2.72} & \scalebox{0.92}{{2.94}}  & \scalebox{0.92}{6.13} & \scalebox{0.92}{3.23} &\scalebox{0.92}{2.89} &\scalebox{0.92}{6.83}&\scalebox{0.92}{3.11}\\
    \multicolumn{1}{c|}{\scalebox{0.92}{11 h}} &  \scalebox{0.92}{\textbf{2.69}} &\scalebox{0.92}{3.08} &\scalebox{0.92}{2.83} & \scalebox{0.92}{{3.16}} & \scalebox{0.92}{7.66} & \scalebox{0.92}{3.06}&\scalebox{0.92}{2.96}&\scalebox{0.92}{7.68}&\scalebox{0.92}{3.14} \\
    \midrule
     \scalebox{0.92}{MAE} & \scalebox{0.92}{\textbf{2.31}} &\scalebox{0.92}{3.27}& \scalebox{0.92}{2.52} &\scalebox{0.92}{{2.79}} & \scalebox{0.92}{4.86}  & \scalebox{0.92}{2.74}&\scalebox{0.92}{2.62}&\scalebox{0.92}{5.47}&\scalebox{0.92}{2.96}\\
    \bottomrule
  \end{tabular}}
  \end{small}
\end{table}

The results on long-term \texttt{Electricity} benchmark is shown in Tab. \ref{tab_elec}, and other two large-scale datasets in Tabs. \ref{tab_cer} and \ref{tab:pv_results}. Although advanced STGNNs achieve competitive performance in medium-sized traffic data, many of them run out of memory on the large-scale energy dataset. Similarly, \gls{model} consistently achieves superior performances under various settings. It is worth commenting that \gls{model} is the only method in the literature related to STGNNs and LTSF that performs well on all of these tasks.

\subsection{Results on Environment Benchmarks}
Environmental records such as weather and air quality contain complicated spatiotemporal processes that are challenging to forecast.
% In addition to traffic data, \gls{model} has the potential to model and analyze general multivariate time series from other domains. 
We further evaluate models' generality using air quality and meteorological data. Results in Tabs. \ref{tab:result_weather} and \ref{tab:air_results} indicate the great potential of \gls{model} to handle time series with complex spatial structures and missing data. Even without popular sequential modeling techniques or advanced spatial models, it still achieves desirable results.

\begin{table}[!htbp]
\caption{Results on \texttt{Weather} (in: 48, out: 24).} 
% Results of baselines are from \cite{wu2023interpretable}.}
\small
\setlength{\tabcolsep}{8pt}
\begin{center}
\resizebox{0.8\columnwidth}{!}{
\begin{tabular}{l | cc| cc }
% \cmidrule[1pt]{2-5}
\toprule
 \multicolumn{1}{c}{\textsc{Datasets}} & \multicolumn{2}{c|}{\texttt{Wind}} & \multicolumn{2}{c}{\texttt{Temp}}  \\
% \cmidrule[0.5pt]{2-5}
\midrule
   \multicolumn{1}{c|}{\textsc{Metrics}} & \scriptsize MSE & \scriptsize MAE & \scriptsize MSE & \scriptsize MAE  \\
 \midrule
% {DeepAR} &  5.25 & 1.60 & 32.25 & 4.26  \\
{Longformer} & 4.47 & 1.49 &17.35 & 3.14    \\
% {FNet} & 4.73 &1.48 & 10.10 &2.24    \\
{Informer} &  4.93 & 1.58 & 33.29 &4.42   \\
{Autoformer} & 4.69 & 1.47 & 10.14 & 2.25    \\
{Prayformer} & 4.61 & 1.51 &  23.33 & 3.67  \\
{FEDformer} & 4.75 & 1.50 & 11.05 &2.41  \\
{ETSformer} & 5.09 & 1.59 & 9.76 & 2.27    \\
% {N-HiTS} & 4.03 & 1.38 & 9.35 & 2.09    \\
% {FreDo} &4.31 &1.43& 14.67 &2.79     \\
{StemGNN} &4.07 & 1.39 & 13.93 & 2.75   \\
{Corrformer} &3.89 & 1.30 & 7.71 & 1.89     \\
\midrule
{N-BEATS} & 4.12 & 1.39 & 9.20 & 2.12     \\
{N-HiTS} & 4.03 & 1.38 & 9.35 & 2.09    \\
\textbf{\gls{model}} & \textbf{3.52}  & \textbf{1.29}  & \textbf{7.43}  &   \textbf{1.87}   \\
\bottomrule
\end{tabular}}
\label{tab:result_weather}
\end{center}
\end{table}

\begin{table}
\caption{Results on \texttt{AQI} benchmark (in: 24, out: 3).}
\small
\setlength{\tabcolsep}{8pt}
\begin{center}
\resizebox{0.8\columnwidth}{!}{
\begin{tabular}{l | ccc | c }
% \cmidrule[1pt]{2-5}
\toprule
\multicolumn{1}{c}{\textsc{Dataset}} & \multicolumn{4}{c}{\texttt{AQI}}   \\
% \cmidrule[0.8pt]{2-5}
\midrule
\multicolumn{1}{c|}{\multirow{2}{*}[-0.45em]{\textsc{Metric}}} & 1 h & 2 h & 3 h & \multicolumn{1}{c}{Ave.} \\
\cmidrule[0.5pt]{2-5}
% \midrule
   \multicolumn{1}{c|}{} & \scriptsize MAE & \scriptsize MAE & \scriptsize MAE & \scriptsize MAE  \\
 \midrule
{AGCRN} & 10.78  & 14.32  & 16.96  &  14.02   \\
 {DCRNN} & 9.89  & 13.79  & 16.78  & 13.49   \\
{GWNet} & \textbf{9.39}  & 12.98  & 15.71  & 12.70  \\
{GatedGN} & 10.39  & 14.68  & 17.76  & 14.28  \\
% {GRUGCN} & 10.02  & 13.95  & 16.92  & 13.63   \\
% {EvolveGCN} & 10.64  & 15.02  & 18.27  & 14.64 \\
% {ST-Transformer} & 10.13  & 14.17  & 17.01  & 13.77   \\
{STGCN} & 9.63  & 13.47  & 16.47  & 13.19 \\
{MTGNN} & 10.24  & 14.10  & 16.83 &  13.72  \\
{D2STGNN} & 9.48  & 12.91  & \textbf{15.44}  &  12.61  \\
% {DGCRN} & -  &  -  & -   &  -  \\
% {DLinear} & 10.99  &  15.79  & 19.31  &  15.36  \\
% {TimesNet} & 17.64  &  19.47  & 21.10  &  19.40 \\
\midrule
{STID} & 10.44  & 14.76 & 17.88  & 14.36  \\
{FreTS} & 12.48 & 17.17 & 20.11 & 16.59     \\
\textbf{\gls{model}} & 9.43 & \textbf{12.90 } & 15.45 & \textbf{12.58} \\
\bottomrule
\end{tabular}}
\label{tab:air_results}
\end{center}
\end{table}

{\subsection{Forecasting result visualization}
As the real-time traffic information is essential for traffic control, especially in congested regions, we provide several examples of speed prediction results using the METR-LA dataset. As indicated by Fig. \ref{fig:speed_viz}, our model can provide accurate predictions for traffic congestions in urban areas.

\begin{figure}[!htbp]
\centering
\subfigure[Speed prediction of sensor \# 2]{
\centering
\includegraphics[width=0.48\columnwidth]{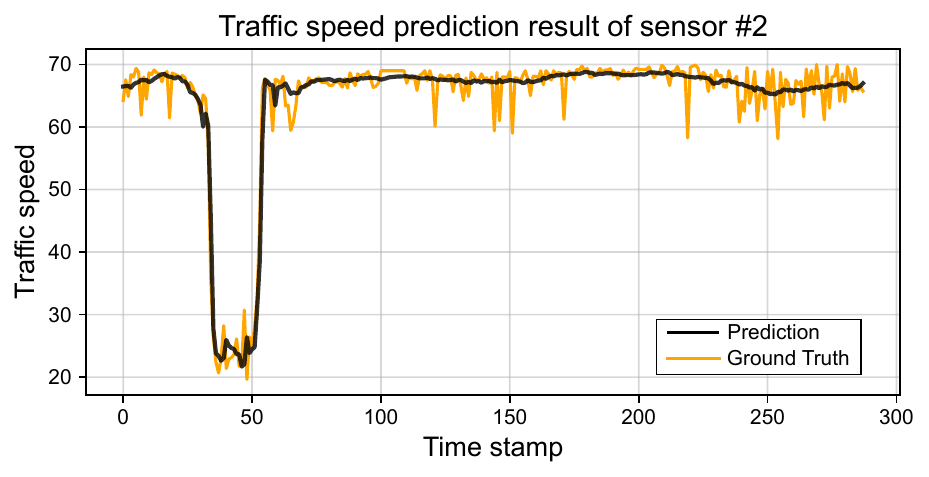}}
\subfigure[Speed prediction of sensor \# 4]{
\centering
\includegraphics[width=0.48\columnwidth]{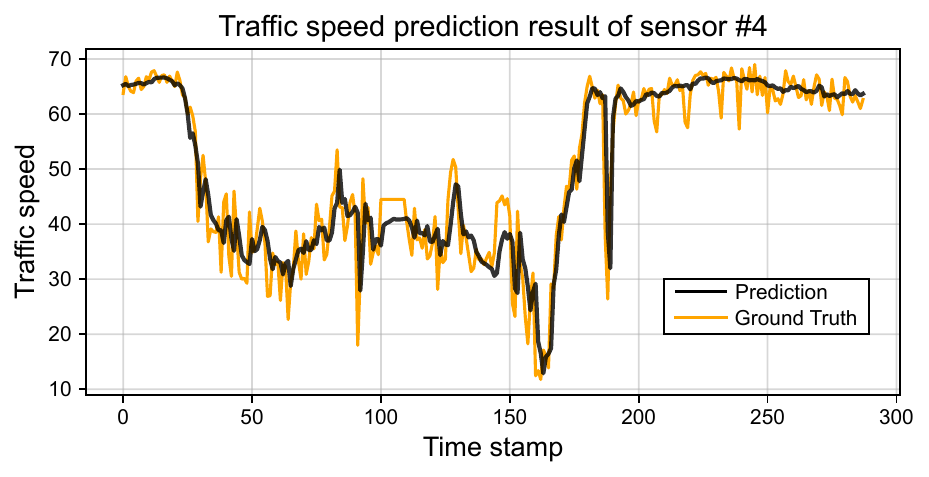}}
\centering
\subfigure[Speed prediction of sensor \# 104]{
\centering
\includegraphics[width=0.48\columnwidth]{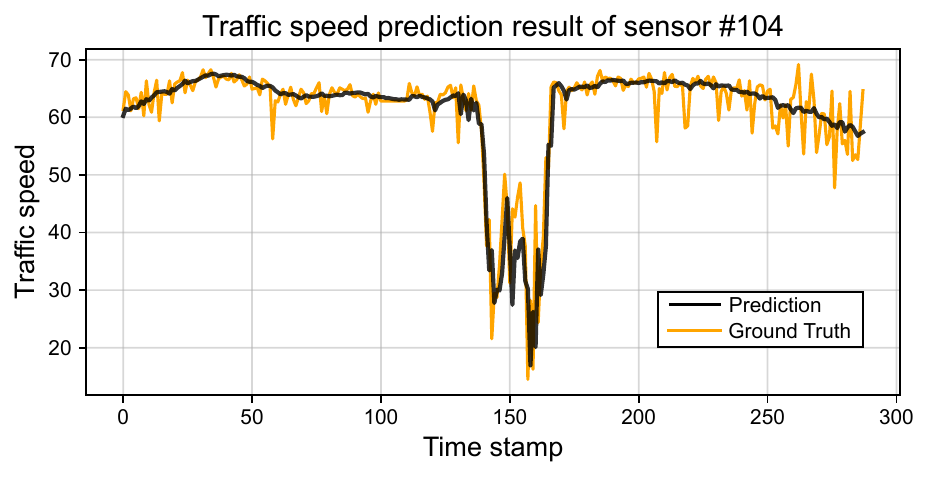}}
\centering
\subfigure[Speed prediction of sensor \# 2]{
\centering
\includegraphics[width=0.48\columnwidth]{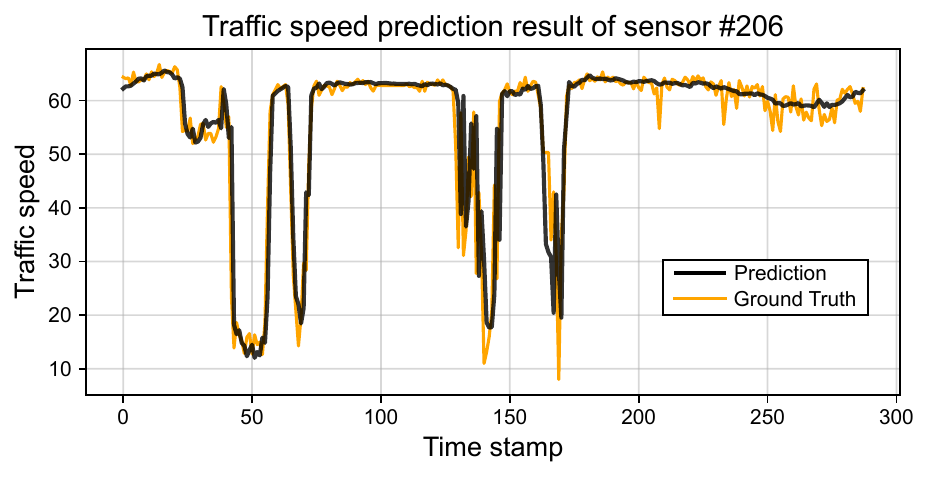}}
\caption{{Traffic speed prediction in congested regions of METR-LA data.}}
\label{fig:speed_viz}
\end{figure}

}

\subsection{Compatibility with probabilistic forecasting}
In practice, the forecasting interval of traffic information is beneficial for quantification of uncertainty.
Our model can be readily extended to generate probabilistic predictions by adopting a compatible loss, e.g., tilted loss (also known as pinball loss) \cite{rodrigues2020beyond}. The deep neural network can then be trained to generate both the mean and quantiles by minimizing the combination of $\ell_1$ loss and tilted loss. To this end, we perform a probabilistic forecasting task and adopt a probabilistic metric, the Continuous Ranked Probability Score (CRPS), to report the uncertainty. Results are shown in Tab. \ref{tab:crps}

\begin{table}[!htbp]
\caption{{Forecasting uncertainty quantification results of PEMS datasets}.}
\small
\begin{center}
\setlength{\tabcolsep}{5pt}
\resizebox{0.8\columnwidth}{!}{
\begin{tabular}{l | ccc | c}
% \cmidrule[1pt]{2-6}
\toprule
 \multicolumn{1}{c}{\textsc{Metric}} & \multicolumn{4}{c}{{CRPS}}  \\
\midrule
 \multicolumn{1}{c|}{{\textsc{Dataset}}} & 3 & 6 & 12 & \multicolumn{1}{c}{Ave.}  \\
 \midrule
{PEMS03} & 0.0744 &  0.0767  & 0.0872 &   0.0775   \\
{PEMS04}  &  0.0720  & 0.0785  & 0.0888  & 0.0810   \\
{PEMS07}  & 0.0823 & 0.0899  & 0.1070 &  0.0909  \\
{PEMS08}  & 0.0499  &  0.0563   & 0.0688  &  0.0566    \\
\bottomrule
\end{tabular}}
\label{tab:crps}
\end{center}
\end{table}

\subsection{Impact of objective metric} We evaluate the impact of objective function in Tab. \ref{tab_metric}.

\begin{table}[!htbp]
\caption{{Impact of different objective metrics}}\label{tab_metric}
  \centering
  \begin{small}
  \renewcommand{\multirowsetup}{\centering}
  \setlength{\tabcolsep}{5pt}
 \resizebox{0.8\columnwidth}{!}{
  \begin{tabular}{c|c| c c |c c }
    % \cmidrule[1pt]{2-7}
    \toprule
     % \multicolumn{2}{c}{\textsc{Dataset}} & \multicolumn{4}{c}{\texttt{Electricity}}  \\
     % \midrule
    \multicolumn{2}{c|}{\textsc{Dataset}} & \multicolumn{2}{c|}{\scalebox{0.8}{PEMS04}}  & \multicolumn{2}{c}{\scalebox{0.8}{PEMS08}} \\
    % \cmidrule[0.7pt]{2-7}
    \midrule
    \multicolumn{2}{c|}{\textsc{Metric}}  & \scalebox{0.92}{$\ell_1$}  & \scalebox{0.92}{$\ell_2$}  & \scalebox{0.92}{$\ell_1$}  & \scalebox{0.92}{$\ell_2$} \\
    \midrule
    \multirow{3}{*}{\rotatebox{90}{MAE}}& \multicolumn{1}{c|}{\scalebox{0.92}{3}} & \scalebox{0.92}{\textbf{17.36}} & \scalebox{0.92}{{17.82}} & \scalebox{0.92}{\textbf{13.15}} & \scalebox{0.92}{13.69}\\
    & \multicolumn{1}{c|}{\scalebox{0.92}{6}} &  \scalebox{0.92}{\textbf{18.04}} & \scalebox{0.92}{{18.64}}  & \scalebox{0.92}{\textbf{13.99}} & \scalebox{0.92}{14.57}\\
    & \multicolumn{1}{c|}{\scalebox{0.92}{12}} &  \scalebox{0.92}{\textbf{19.13}} & \scalebox{0.92}{{19.84}} & \scalebox{0.92}{\textbf{15.26}} & \scalebox{0.92}{ 16.05}\\
    \midrule
     \multirow{3}{*}{\rotatebox{90}{Ave.}} &\scalebox{0.92}{MAE} & \scalebox{0.92}{\textbf{18.03}} & \scalebox{0.92}{{18.59}} & \scalebox{0.92}{\textbf{13.99}}  & \scalebox{0.92}{14.60}\\
    & \scalebox{0.92}{MSE} & \scalebox{0.92}{\textbf{871.34}} & \scalebox{0.92}{{921.15}} & \scalebox{0.92}{\textbf{533.09}} & \scalebox{0.92}{560.87}\\
     & \scalebox{0.92}{MAPE} & \scalebox{0.92}{\textbf{12.48\%}} & \scalebox{0.92}{{13.00\%}} & \scalebox{0.92}{\textbf{9.05\%}}  & \scalebox{0.92}{9.61\%}\\
    \bottomrule
  \end{tabular}}
  \end{small}
\end{table}

As can be seen, the $\ell_1$ loss function consistently achieves better performances in out studies. This may be due to its robustness to anomalies.

% \subsection{Additional Interpretations}
\subsection{Additional Architecture Details}
This sections provides the detailed implementations of \gls{model}, as a technical complement to the descriptions presented in the paper.
To give a clear exposition of the design concept of our model, we provide more discussions and interpretations on the model architectures.

\paragraph{Time Stamp Encoding}
We adopt sinusoidal positional encoding to inject the time-of-day information along time dimension:
\begin{equation}
    \begin{aligned}
        PE_{\text{sine}}&=\texttt{sin}(p_i*2\pi/\delta_D),\\
        PE_{\text{cosine}}&=\texttt{cos}(p_i*2\pi/\delta_D),\\
        \mathbf{U}_t &= \left[PE_{\text{sine}} \| PE_{\text{cosine}}\right],
    \end{aligned}
\end{equation}
where $p_i$ is the index of $i$-th time point in the series, and $\delta_D$ is the day-unit time mapping. We concatenate $PE_{\text{sine}}$ and $PE_{\text{cosine}}$ as the final temporal encoding. In fact, day-of-week embedding can also be applied using the one-hot encoding.

\paragraph{Node Embedding} 
For the spatial dimension, we adopt a unique identifier for each sensor. While an optional structural embedding is the random-walk diffusion matrix \cite{dwivedi2020benchmarking}, for simplicity, we use the learnable node embedding \cite{shao2022spatial} as a simple index positional encoding without any structural priors. Learnable node embedding can be easily implemented by initializing a parameter with its gradient trackable. The initialization method (e.g., Gaussian or uniform distribution) can be used to specify its initial distribution.

\paragraph{Dense Readout}
For a multi-step \acrshort{task} task, we adopt a $\textsc{Mlp}$ and a reshaping layer to directly output the predictions:
\begin{equation}
\begin{aligned}
    &\widehat{\mathbf{X}}_{T+1:T+H}=\textsc{Mlp}(\mathbf{H}^{(L+1)}),\\
    &\widehat{\mathcal{X}}_{T+1:T+H}=\texttt{UNFOLD}(\widehat{\mathbf{X}}_{T+1:T+H}),
\end{aligned}
\end{equation}
\noindent where $\texttt{UNFOLD}(\cdot)$ is the inverse linear operator of $\texttt{FOLD}(\cdot)$. For simplicity, we avoid a complex sequential decoder and directly make multi-step predictions through regression.

\paragraph{Interpretations of the Temporal Contextualization Issue}
Different from the spatial contextualization issue, the temporal contextualization issue can be shown in a typical linear predictive model with weight $\mathbf{W}$ and bias $\mathbf{b}$, expressed as follows:
\begin{equation}\label{autoreg}
\begin{aligned}
    &\mathbf{x}_{t+1:t+H}=\mathbf{W}\mathbf{x}_{t-W:t}+\mathbf{b},\\
    \text{or:~}&x_{t+h}=\sum_{k=0}^W w_{k,h}x_{t-k}+b_{k,h},~h\in\{1,\dots,H\}.
\end{aligned}
\end{equation}
Eq. \eqref{autoreg} is an autoregressive model (AR) for each forecast horizon. Its weight $w_{k,h}$ depends solely on the relative time order and is agnostic to the absolute position in the sequence, rendering it incapable of contextualizing the series in temporal dimension. In this case, the temporal context is needed.

In the discussion by \cite{chen2023tsmixer}, predictive models like Eq. \eqref{autoreg} are referred to as time-dependent. An upgrade to this type is termed data-dependent, where the weight becomes pattern-aware and conditions on temporal variations:
\begin{equation}\label{time_varying_ar}
x_{t+h}=\sum_{k=0}^W\mathcal{F}_k(\mathbf{x}_{t-W:t})x_{t-k}+b_{k,h},
\end{equation}
\noindent where $\mathcal{F}_k(\cdot)$ represents a data-driven function, such as self-attention. Eq. \eqref{time_varying_ar} creates a fully time-varying AR process, which is parameterized by time-varying coefficients \cite{bringmann2017changing}. However, its overparameterization can lead to overfitting of the data, rather than capturing the temporal relationships, such as the position on the time axis. 
\bibliographystyle{IEEEtran}
\bibliography{references}
% \vfill

% \bf{If you include a photo:}\vspace{-33pt}
% \begin{IEEEbiography}[{\includegraphics[width=1in,height=1.25in,clip,keepaspectratio]{fig1}}]{Michael Shell}
% Use $\backslash${\tt{begin\{IEEEbiography\}}} and then for the 1st argument use $\backslash${\tt{includegraphics}} to declare and link the author photo.
% Use the author name as the 3rd argument followed by the biography text.
% \end{IEEEbiography}
% \vspace{11pt}

\begin{IEEEbiography}[{\includegraphics[width=1in,height=1.25in,keepaspectratio]{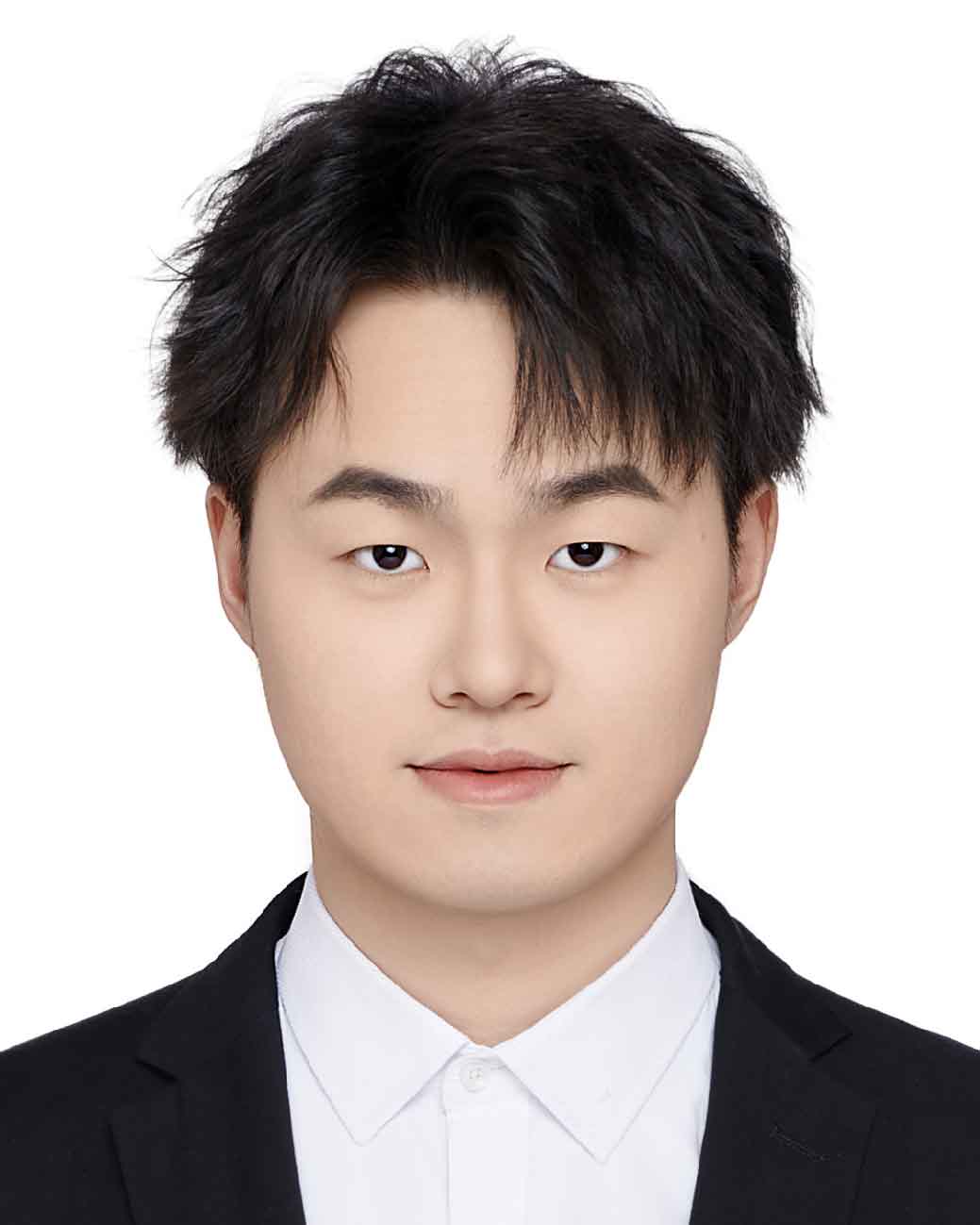}}]{Tong Nie (Student Member, IEEE)} received the B.S. degree from the college of civil engineering, Tongji University, Shanghai, China, in 2022. He is currently pursuing dual Ph.D. degrees with the Department of Traffic Engineering in Tongji University and the Department of Civil and Environmental Engineering in The Hong Kong Polytechnic University. He has published several papers in top-tier journals and conferences, including SIGKDD, CIKM, TR-Part C, IEEE T-ITS, and IEEE T-IV.
His research interests include spatiotemporal data modeling, time series analysis, and graph-based deep learning.
\end{IEEEbiography}
% \vspace{11pt}

\begin{IEEEbiography}[{\includegraphics[width=1in,height=1.25in,keepaspectratio]{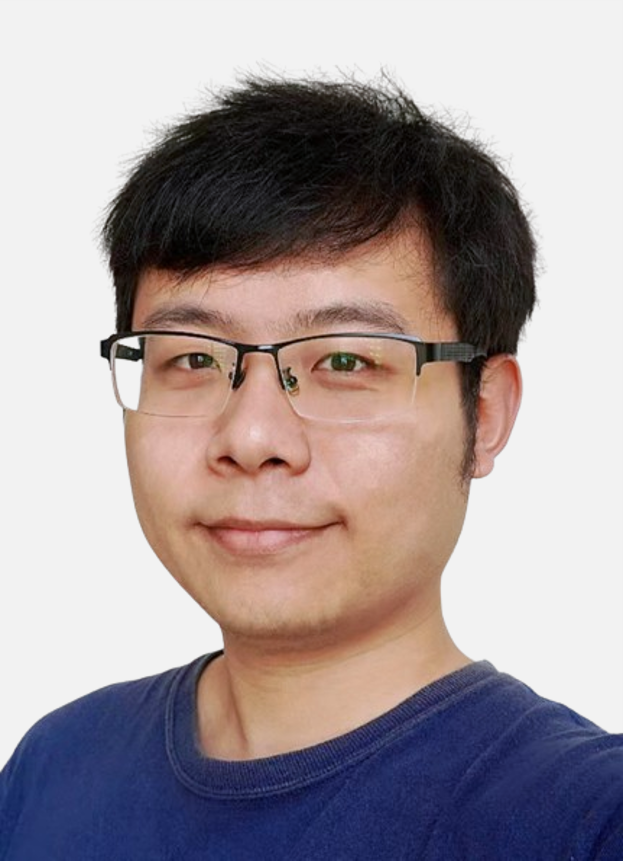}}]{Guoyang Qin} received his Ph.D. in transportation engineering from Tongji University in Shanghai, China, in 2021. Currently, he works as a postdoctoral researcher at Tongji University. He has published 8 papers in Transportation Research Part C, a renowned journal in the field of transportation research. His research is centered around understanding the complexity and patterns of urban mobility systems through modeling, data science, and visualization. Based on this understanding, he aims to optimize the systems by employing techniques such as deep learning, reinforcement learning, and privacy preservation.
\end{IEEEbiography}

\begin{IEEEbiography}[{\includegraphics[width=1in,height=1.25in,keepaspectratio]{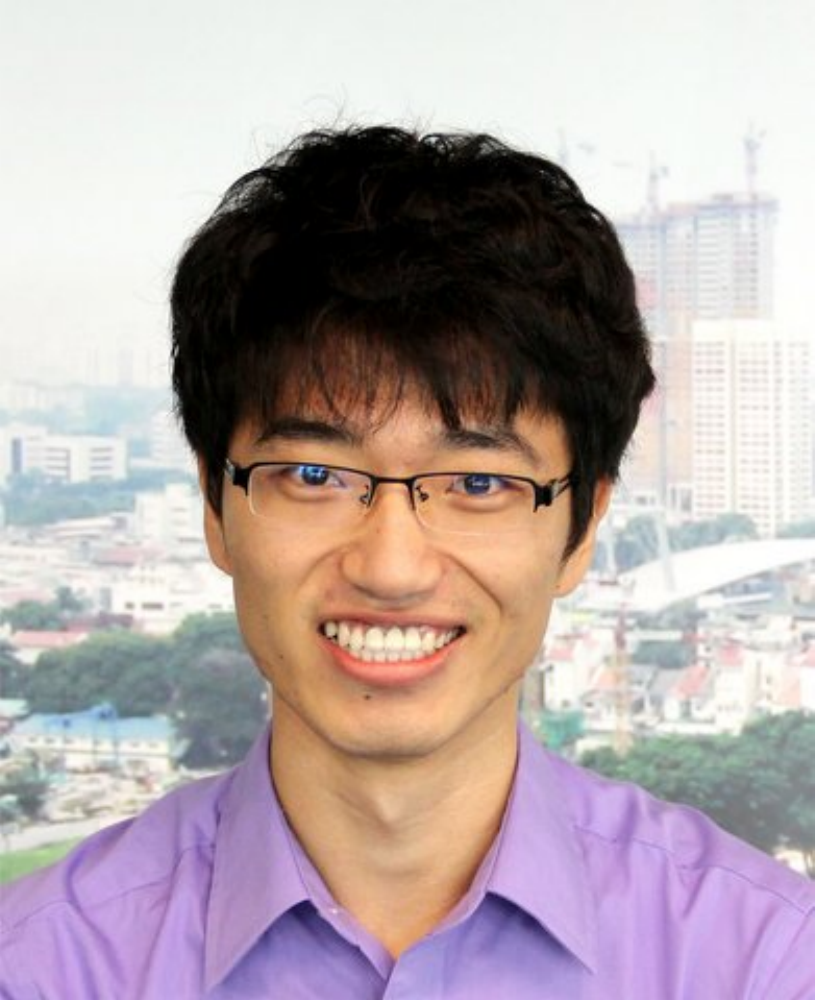}}]{Lijun Sun (Senior Member, IEEE)} received the
B.S. degree in civil engineering from Tsinghua University, Beijing, China, in 2011, and the Ph.D. degree in civil engineering (transportation) from the National the University of Singapore
in 2015. He is currently an Associate Professor and William Dawson Scholar in the Department of Civil Engineering, McGill University, Montreal, QC, Canada. His research centers on intelligent transportation systems, traffic control and management, spatiotemporal modeling,
Bayesian statistics, and agent-based simulation.
\end{IEEEbiography}

\begin{IEEEbiography}
[{\includegraphics[width=1in,height=1.25in,keepaspectratio]{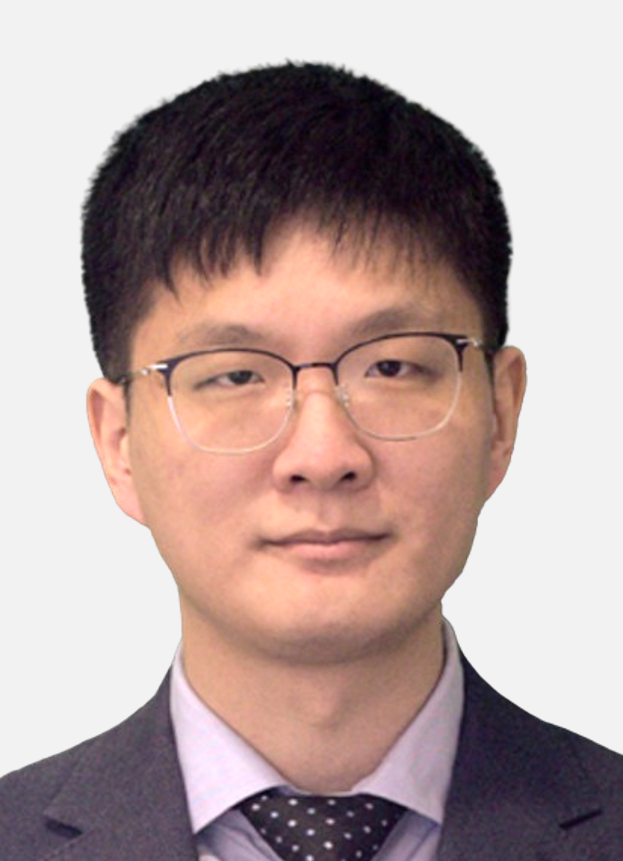}}]{Wei Ma (Member, IEEE)} received the bachelor’s
degree in civil engineering and mathematics from
Tsinghua University, China, and the master’s degree
in machine learning and civil and environmental engineering and the Ph.D. degree in civil and
environmental engineering from Carnegie Mellon
University, USA. He is currently an Assistant Professor with the Department of Civil and Environmental
Engineering, The Hong Kong Polytechnic University (PolyU). His current research interests include
machine learning, data mining, and transportation
network modeling, with applications for smart and sustainable mobility
systems.
\end{IEEEbiography}

% \vfill

\begin{IEEEbiography}[{\includegraphics[width=1in,height=1.25in,keepaspectratio]{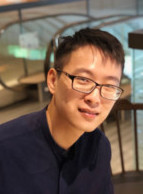}}]{Yu Mei} received the M.S. degree in transportation engineering from Tongji University, Shanghai, China, in 2015, and the Ph.D degree in transportation from The Hong Kong Polytechnic University in 2019.
Now, he is the team leader of the Department of Intelligent Transportation Systems, Baidu Inc., Beijing, China.
His research interests include intelligent transportation systems, traffic signal control, and autonomous driving navigation.
\end{IEEEbiography}

% \vspace{-6.5ex} 

\begin{IEEEbiography}[{\includegraphics[width=1in,height=1.25in,keepaspectratio]{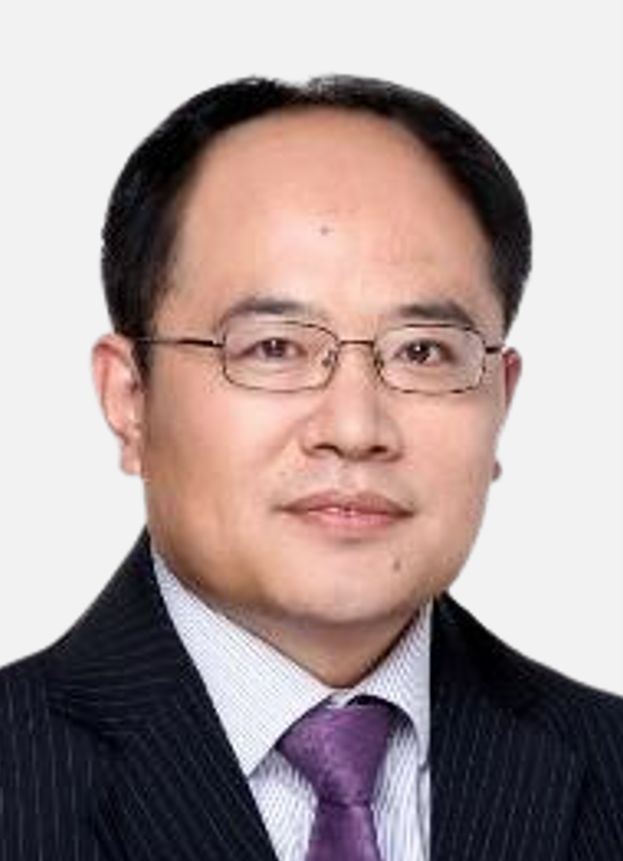}}]{Jian Sun} received the Ph.D. degree in transportation engineering from Tongji University, Shanghai, China. He is currently a Professor of transportation engineering with Tongji University. He has published more than 200 papers in SCI journals.
His research interests include intelligent transportation systems, traffic flow theory, AI in transportation, and traffic simulation. 
\end{IEEEbiography}
% \vspace{11pt}
\vfill

\end{document}